\documentclass[10pt,twocolumn,letterpaper]{article}

\usepackage[pagenumbers]{cvpr}

\usepackage{graphicx}
\usepackage{amsmath}
\usepackage{amssymb}
\usepackage{booktabs}

\usepackage{epsfig}

\usepackage{graphicx}

\usepackage{verbatim}

\usepackage{float}
\DeclareMathOperator{\sign}{\it{Sign}}

\usepackage[pagebackref=true,breaklinks=true,letterpaper=true,colorlinks,bookmarks=false]{hyperref}

\usepackage[capitalize]{cleveref}
\crefname{section}{Sec.}{Secs.}
\Crefname{section}{Section}{Sections}
\Crefname{table}{Table}{Tables}
\crefname{table}{Tab.}{Tabs.}

\begin{document}

\title{Consistent Semantic Attacks on Optical Flow}

\author{Tom Koren~~~~~~~
Lior Talker~~~~~~~
Michael Dinerstein~~~~~~~ 
Roy J Jevnisek \\
Samsung Israel R\&D Center, Tel Aviv, Israel \\
{\tt\small tomkore@gmail.com};
{\tt\small \{lior.talker, m.dinerstein, roy.jewnisek\}@samsung.com}
}

\maketitle

\begin{abstract}

We present a novel approach for semantically targeted adversarial attacks on Optical Flow. In such attacks the goal is to corrupt the flow predictions of a specific object category or instance. Usually, an attacker seeks to hide the adversarial perturbations in the input. However, a quick scan of the output reveals the attack. In contrast, our method helps to hide the attacker’s intent in the output as well. We achieve this thanks to a regularization term that encourages off-target consistency. We perform extensive tests on leading optical flow models to demonstrate the benefits of our approach in both white-box and black-box
settings. Also, we demonstrate the effectiveness of our attack on subsequent tasks that depend on the optical flow.

\end{abstract}



\section{Introduction}
\label{sec:intro}

Optical Flow (OF) is a crucial subtask of many safety-critical pipelines. It is especially important for Advanced Driver Assistance Systems (ADAS) and autonomous vehicles,
where unreliable optical flow can be hazardous and life-threatening. For example, \textit{Time-To-Collision} (TTC) methods often rely on optical flow \cite{yang2020upgrading, pedro2021collision, blumenkamp2019end, badki2021binary}, and their errors can have dangerous consequences.

In this paper we consider malicious manipulations aiming to lead the OF predictions astray.
These manipulations are represented by perturbations, sometimes subtle, that are introduced into the input pixels.
In the literature such perturbations are referred to as Adversarial Attacks (AA) \cite{KurakinGB16a,FGSM_goodfellow2014,szegedy2013intriguing}.

\begin{figure}[t]
\begin{center}
   \includegraphics[width=1.0\linewidth]{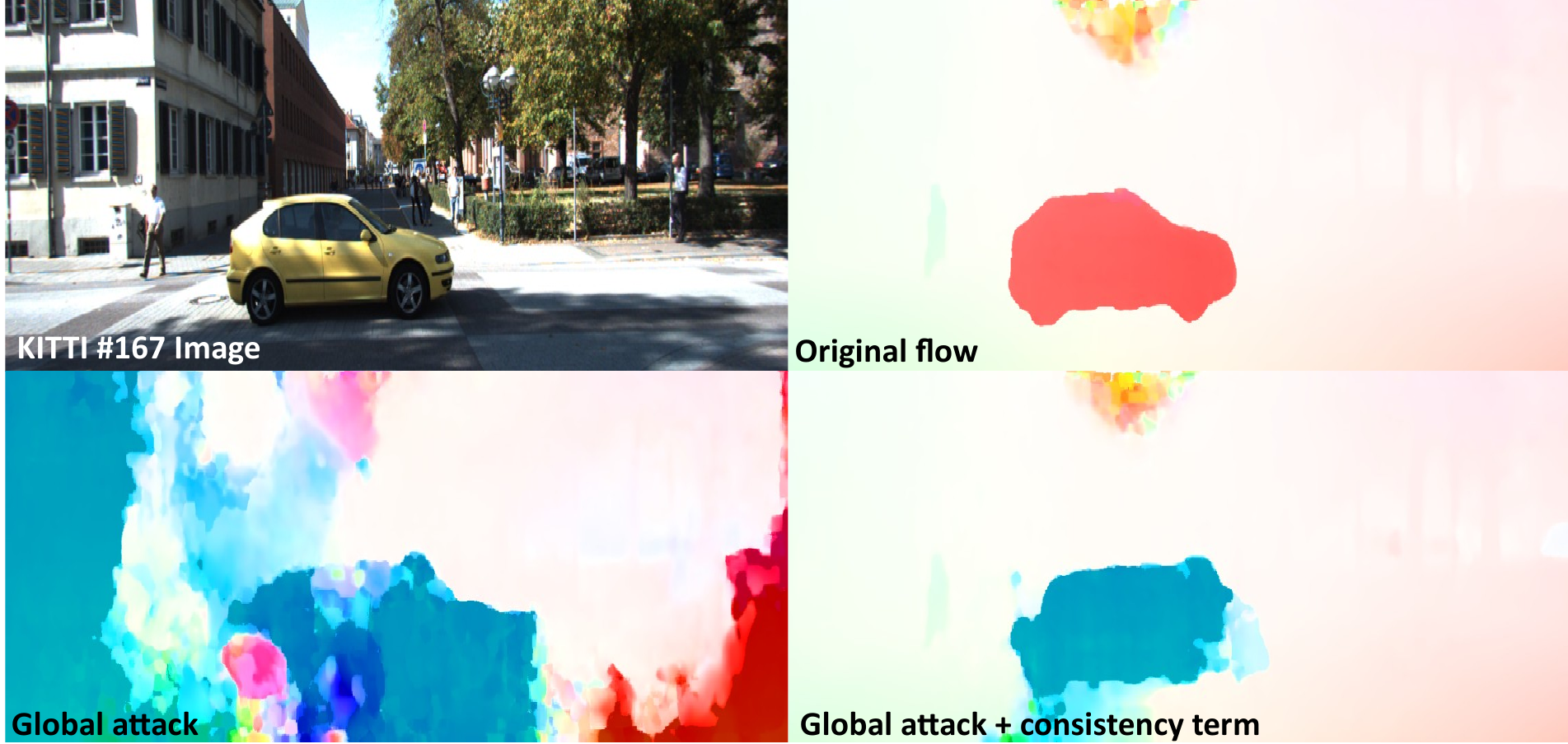} 
   \vspace{-0.5cm}
\end{center}
   \caption{An example of a targeted attack on HD3-PPAC \cite{PPAC_20} optical flow model. Top: the original predicted flow and the corresponding input image. Bottom left: the corrupted flow using a non-consistent attack. Bottom right: the corrupted flow using a consistent targeted attack on the vehicle in the scene.}
\vspace{-0.5cm}
\label{fig:Figure1}
\label{fig:onecol}
\end{figure}

The attacker's goal is to damage a system's performance and remain unnoticed. The defender's goal is to design a system that operates reliably despite such attacks.
 To achieve this goal, the defender can, for example, use some AA detection method to discard suspicious inputs.
One approach to detect AAs\cite{grosse2017statistical, tian2018detecting}, is to examine the input to the attacked model. Another approach, which we consider in this paper, is to examine the \emph{output} of the attacked model. We show that a straightforward attack on OF may be fairly easy to detect in the output. 
 We propose an AA method, which is more difficult to detect, but has a similar or stronger effect on OF.  

We use the following observation: in the context of automotive applications some objects are more important than others.
Obvious examples of such important objects are pedestrians and vehicles.
For TTC systems, failing to estimate the correct flow for pedestrians and vehicles can lead to fatal accidents.

Assuming a semantic or instance segmentation of the observed scene is given, an attacker can specify a target to attack. 
That is, instead of targeting the entire image, only a subset, defined by its semantic segmentation, can be selected.
Likewise, instead of perturbing all the pixels in the image, a subset of the pixels can be chosen at which the attacker introduces malicious perturbations.

Adversarial attacks targeting only a subset of pixels may still alter the predictions of other pixels in the image. To make the attack less detectable, it is beneficial to corrupt the target prediction without affecting the rest of the predictions. We refer to attacks that leave off-target predictions unaffected (or affected as little as possible) as "consistent attacks". In this paper we present a method to create targeted consistent adversarial attacks on optical flow. The chosen targets for the attacks are the "vehicle" and the "human" categories.
 
Figure \ref{fig:Figure1} depicts an example of such an attack.
The bottom row presents two attacked flows (encoded using Middlebury \cite{Middlebury_BakerSLRBS11} color-wheel): consistent and non-consistent. 
Both attacks achieve their goal - the flow of the corresponding target, the vehicle, is heavily damaged.
However, the difference between them is readily seen. 
A consistently attacked flow looks reasonable, while a non-consistent attack results in a flow which is chaotically cluttered. In this paper we show that the first attack is less detectable than the second.

To achieve the effect described above, we introduce a new optimization term. 
We refer to it as "consistency term". 
While being a relatively simple addition to the optimization loss, the consistency term leads to multiple improvements in the generated adversarial attacks when compared to the baseline non-consistent settings. 
First, as expected, the impact on the flow predictions of non-target scene objects is significantly reduced. 
Second, the effectiveness of attacks on targets increases.
Third, our experiments show that in the "black-box" setting, i.e. where the attacked model is inaccessible, we observe a much better transferability \cite{papernot2017practical}.
Finally, we demonstrate that consistent attacks are more effective on a TTC system while being less noticeable by detection methods.

We have conducted an extensive set of evaluations using five leading optical flow methods \cite{LFN_18,RAFT_20,PPAC_20,VCN_19,HD3_19}.
There are three main groups of experiments where we compare attacks obtained with and without our consistency term: global, local and cross-category attacks. 
The target of those attacks is always the same, but the subsets of pixels that are perturbed are different.
In a global attack setting, the perturbation can be distributed over all pixels. 
In a local attack setting, only pixels of the target object can be perturbed. 
Finally, in a cross-category setting, in order to corrupt the flow of some target object, we perturb pixels of some other object. 
We have also evaluated the effect of these attacks on a downstream TTC task. We compare consistent and non-consistent attacks in terms of the tradeoff between their effect on TTC and a AA detection score.

To summarize our contribution, we are the first to study targeted attacks on optical flow models.
We introduce a new term to the optimization loss, which we name the "consistency" term, to preserve the optical flow of non-target objects. 
This helps to hide the attacks in the output of the optical flow.
We show that the resulting consistent attacks are more effective than the non-consistent attacks. 
We demonstrate that these attacks are more transferable, i.e. more efficient in a black-box setting, than the non-consistent attacks. 
Finally, we show that under three detection methods these attacks are more effective against a downstream TTC system.

\subsection{Related Work}
The history of optical flow methods goes back to the early 1980s, when the foundational studies of Lucas-Kanade \cite{Lucas_Kanade81} and Horn-Schunk \cite{Horn_Schunck81} were published. 
Since then, hundreds of classical computer vision techniques were proposed.
A very substantive survey on the non-deep optical flow methods can be found in \cite{BlackOF_Survey_14}.
In the new era of deep learning models in computer vision, much attention has been paid to optical flow. 
Deep neural network (DNN) based optical flow methods such as \cite{FlowNet_DosovitskiyFIHH15, FlowNet2_IlgMSKDB17,LFN_18,PWC_SunY0K18,RAFT_20,PPAC_20,VCN_19,HD3_19} have left  the classical approaches far behind in terms of performance. 
This is clearly seen in the results of the KITTI'15 benchmark \cite{KITTI_15}, where the leading non-deep optical flow method \cite{Hu_2017_CVPR} is scored about 150-th place.

DNN based optical flow models can be divided into groups according to their architecture characteristics: encoder-decoder \cite{FlowNet_DosovitskiyFIHH15, FlowNet2_IlgMSKDB17} and spatial pyramid \cite{RanjanB17,LFN_18,PWC_SunY0K18,HD3_19,PPAC_20} networks. 
Some models (\cite {PWC_SunY0K18, HD3_19, PPAC_20, VCN_19}) use a coarse-to-fine technique to refine their predictions. 
Others \cite {RAFT_20} operate with full resolution features at every stage of the model.
In addition, a model can be equipped with a recurrent refinement mechanism, which is placed on top of an optical flow model, as in \cite{IRR_Hur019}. 
Finally, the RAFT model \cite{RAFT_20}, which has demonstrated the \textit{state-of-the-art} performance on KITTI'15 \cite{KITTI_15}, consists of the encoder-decoder part followed by a simple recurrent module utilizing GRU \cite{GRU_Cho2014} blocks.
The use of deep neural networks has opened the door for adversarial perturbations.

Historically, many of the early adversarial attacks were carried out in the context of image classification tasks \cite{szegedy2013intriguing, FGSM_goodfellow2014, earlyattacks_nguyen2015deep, earlyattacks_moosavi2016deepfool}.
The attacker's goal was to force a model to misclassify the input image. 
Many attack schemes were developed and tested on such models. 
One of the most cited is the so-called Fast Gradient Sign Method (FGSM) \cite{FGSM_goodfellow2014}.
In their original work, Goodfellow et al \cite{FGSM_goodfellow2014} suggest a fast method to create adversarial input to a classification model. Consider an input $x$ to a classification model $M$, a hyper-parameter $\epsilon$, a loss function $l$ and $y_{true}$ - the target associated with $x$. 
Assume the model predicts a label y for an input x, i.e., M(x)=y. In their work they show that an adversarial example $x_{adv}$ could then be computed by $x_{adv}=x+\epsilon Sign( \nabla(l(M(x),y_{true}))$. Shortly after \cite{IFGSM_KurakinGB17a} introduces a straightforward way to extend this method. They named this new approach the Iterative Fast Gradient Sign Method (IFGSM) \cite{IFGSM_KurakinGB17a}. They suggest to iteratively use the same update step on the input. To do so, set $x_{adv}^{0}=x$ and iteratively update $x_{adv}^{i+1}=x_{adv}^{i}+\epsilon \cdot Sign( \nabla(l(M(x_{adv}^{i}),y_{true}))$. Since these attacks are thoroughly researched and well understood \cite{understood_tramer2017ensemble, FGSM_goodfellow2014, understood_madry2017towards} we adopt them to our attack approach. 

Later on, adversarial attack methods that target specific objects in the image were introduced against object detectors \cite{DetectAA_20, zhang2019towards, liao2020fast}. 
In \cite{DetectAA_20} it is shown how to force a SOTA detection model to classify all detections of a semantic class as another class while leaving all other detections unchanged.
Liao \textit{et al} \cite{ liao2020fast} proposed a local attack that only perturb a specific detection bounding box, achieving a stronger effect than a global perturbation for the same attack budget.
In \cite{zhang2019towards}, an analysis of object detection from the viewpoint of multi-task learning leads to a method to (partially) defend object detectors against adversarial attacks.

Recently, adversarial attacks have expanded beyond image classification and object detection to include dense prediction tasks such as semantic segmentation, depth, and optical flow.
Promising results are shown in each of these tasks \cite{SemSegAA_FischerKMB17,SemSegAA_ArnabMT18,Geiger_OF,wong2020targeted}.
Such attacks often demonstrate the ability to target specific subsets of pixels rather than the entire image.
For semantic segmentation, it was shown that pixels belonging to specific instances of pedestrians can be labeled by the attacked model as a road \cite{SemSegAA_FischerKMB17}.
In \cite {wong2020targeted}, depth prediction has been successfully manipulated in many ways, such as removing the target entirely and aligning its depth with the surrounding background.

Recently, there is a growing interest in adversarial attacks on optical flow models \cite{Geiger_OF, schrodi2021causes, inkawhich2018adversarial , anand2020adversarial, yamanaka2021simultaneous}.
Ranjan \textit{et al} \cite{Geiger_OF} demonstrated the possible benefits of a patch attack against leading models.
First, they showed that this attack is very successful against encoder-decoder like architectures, but less effective for spatial pyramid types of models.
They also showed it to be reproducible in "real life" conditions, with a hostile patch printed on a board and displayed in front of a camera. 
A follow-up work \cite{schrodi2021causes} conjectures that a principal cause for the success of adversarial attacks on OF is the small size of their receptive field.
Finally, \cite{inkawhich2018adversarial , anand2020adversarial} introduce methods to corrupt the prediction of action recognition systems by attacking the OF modules they rely on. 
Differently from the above, we consider the effectiveness of adversarial attacks on OF methods from the perspective of the ability to hide them in the output, in addition to their impact on performance.

Finally, one of the most important tools in risk assessment and collision avoidance for autonomous agents, e.g., robots and autonomous vehicles, is estimating the TTC \cite{manglik2019future, mori2013first, yang2020upgrading, pedro2021collision, blumenkamp2019end, badki2021binary}. A popular approach to estimate the TTC is using OF \cite{yang2020upgrading, pedro2021collision, blumenkamp2019end, badki2021binary}. For example, \cite{yang2020upgrading} fuses 2D OF vectors, and per-pixel estimated scale change, to "upgrade" 2D OF to 3D, allowing the direct computation of the TTC. We use \cite{yang2020upgrading} to demonstrate the impact of our consistency term on the TTC, and the benefits it has over the non-consistent attack.



\section{Method} 
The inputs for an optical flow network $f_{flow}$ are two $H{\mkern-2mu\times\mkern-2mu} W{\mkern-2mu\times\mkern-2mu} 3$ RGB images $I_1(x,y), I_2(x,y)$ where RGB channels ranges between [0,1]. The output is an $H{\mkern-2mu\times\mkern-2mu} W{\mkern-2mu\times\mkern-2mu} 2$ optical flow vector map $V(x,y)$ .

The goal of an attacker is to find an additive perturbation to the input that would shift the attacked optical flow map $V'$ away from the original prediction $V$, as in \cite{Geiger_OF}.

\begin{figure}[t]
\begin{center}
   \includegraphics[width=1.0\linewidth]{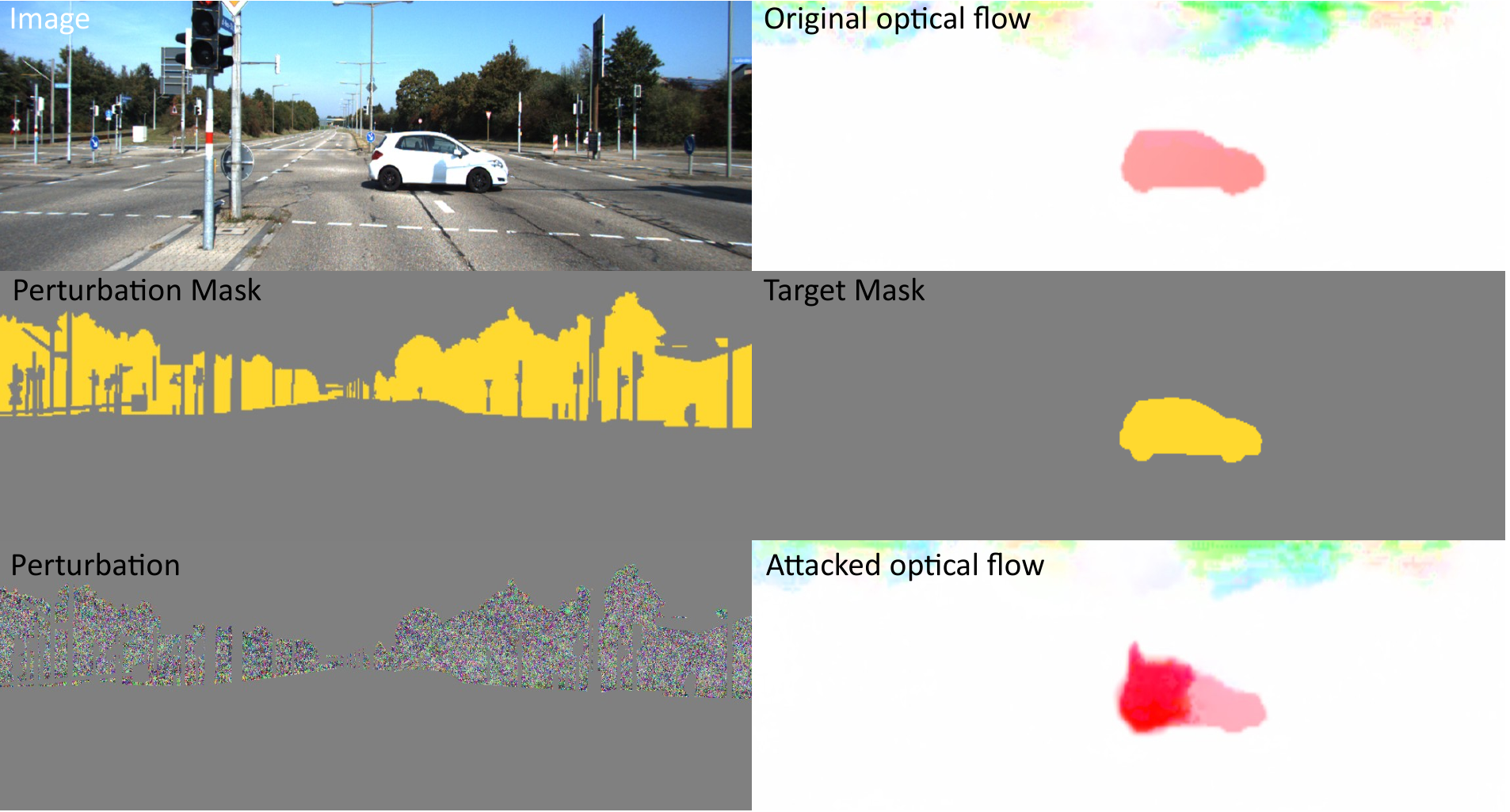} 
   \vspace{-0.6cm}
\end{center}
   \caption{ Consistent attack method. The consistent attack takes as input two masks. The first determines perturbation locations. The second determines the target of the attack. In this example we perturb nature pixels in order to damage the optical flow of a vehicle instance. The top row shows the image and original flow. Middle row shows the perturbation mask, $M_{perturb}$ (left), and the target mask, $M_{target}$ (right). Bottom row shows the perturbation (left) and the resulting attacked optical flow (right).}
\label{fig:FigureMethod}
\vspace{-0.3cm}
\end{figure}

To calculate this perturbation, we use two binary masks. Figure \ref{fig:FigureMethod} visualizes a possible selection of these masks. The first mask, $M_{target}$, selects target pixels. In this example we aim to alter a vehicle's instance flow, and thus $M_{target}$ is the vehicle's instance mask. The second mask, $M_{perturb}$, specifies the pixels we alter. In this example we allow the perturbation to alter only the nature category pixels and so  $M_{perturb}$ is a mask consisting of all nature pixels. 

Consider the first mask $M_{target}$ with $N$ non-zero entries specifying the pixels of the object (category or instance) we aim to attack. Our attack term, $l_{attack}(V',V)$, is then defined by the $L1$ difference between the attacked 
$V'$ and original $V$ flow, averaged only on attacked mask pixels, as given by
\begin{equation}
l_{attack}(V',V) = \frac{1}{N} \sum_{(x,y)\in M} |V'(x,y)-V(x,y)|_1, \label{eq:eq1}
\end{equation}
where $(x,y)\in M$ iff $M_{target} (x,y)=1$.

To encourage the flow on the remaining scene to stay unaffected by the attack, we add a consistency term, $l_{consistency}(V',V)$, which is the negative $L1$ norm of the difference between original and attacked flows, averaged over non-attacked pixels:
\begin{equation}
\begin{split}
&l_{consistency}(V',V)=\\
&-\frac{1}{HW-N} \sum_{(x,y)\notin M} |V'(x,y)-V(x,y)|_1,
\end{split}
\label{eq:eq2}
\end{equation}
where $(x,y)\notin M$ iff $M_{target}(x,y)=0$, and since $N$ pixels are attacked, we have $HW-N$ non-attacked pixels.

Our final loss is composed of these two terms, $l_{attack}$ and $l_{consistency}$. The trade-off between the terms is controlled by the consistency coefficient $\alpha$:
\begin{equation}
l_{total}=l_{attack}+\alpha l_{consistency}
\label{eq:eq3}
\end{equation}

In order to attack semantic categories we require ground truth semantic labeling. This is only provided for the first image $I_1$  in the data we use for the attacks. Thus we have restricted our perturbation to the first image $I_1$. The second image $I_2$ is left unperturbed by our attack. 

Let us denote the first image after the $i$-th perturbation as $I_1^{(i)}$ , the $i$'th perturbation as $\delta I_1^{(i)}$ and the corresponding attacked flow $V^{(i)}$. Thus $V^{(0)}=V$ is the original flow, and $I_1^{(0)}=I_1$ is the unperturbed image. Since our first attack step is when $i=1$ we have $\delta I_1^{(0)}=0$.

Consider a mask, $M_{perturb}$, with $L$ non-zero entries, of the pixels we allow the attack to perturb. 
Given an attack strength coefficient $\epsilon$, our $i$-th attack step follows the IFGSM \cite{IFGSM_KurakinGB17a} and given by 
\begin{figure*}[!tp]
\begin{center}
   \includegraphics[width=1.0\linewidth]{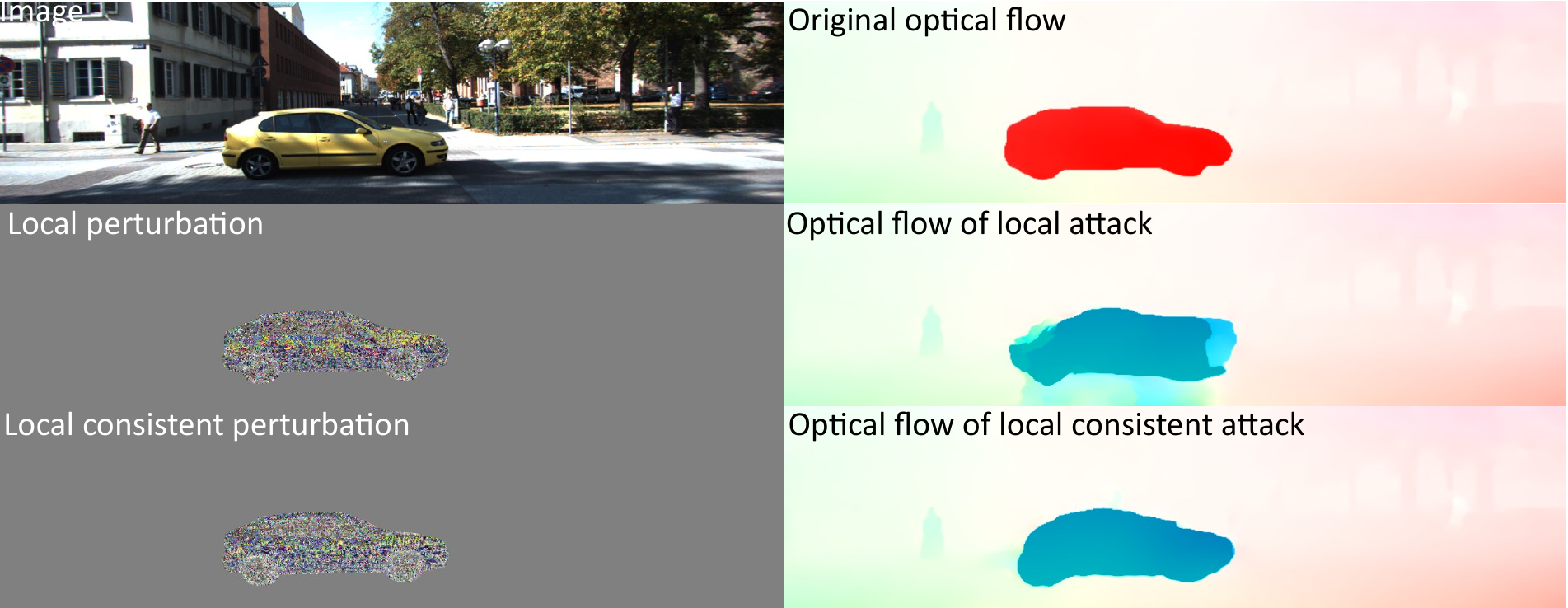} 
   \vspace{-0.6cm}
\end{center}
   \caption{ A visualization of a local attack baseline and the impact of the consistency term on a vehicle instance using LFN \cite{LFN_18} and $||\Delta{I}||=2 \cdot10^{-2}$. Left: the original image and the perturbation optimized by each attack. Right: the corresponding optical flows. Adding a consistency term reduces the effect on non-vehicle pixels (as can be seen below the vehicle), while still significantly changing the vehicle's optical flow.}
   \vspace{-0.5cm}
\label{fig:Figure1Local}
\end{figure*}

\begin{figure}[th!]
\begin{center}
	\includegraphics[width=1.0\linewidth]{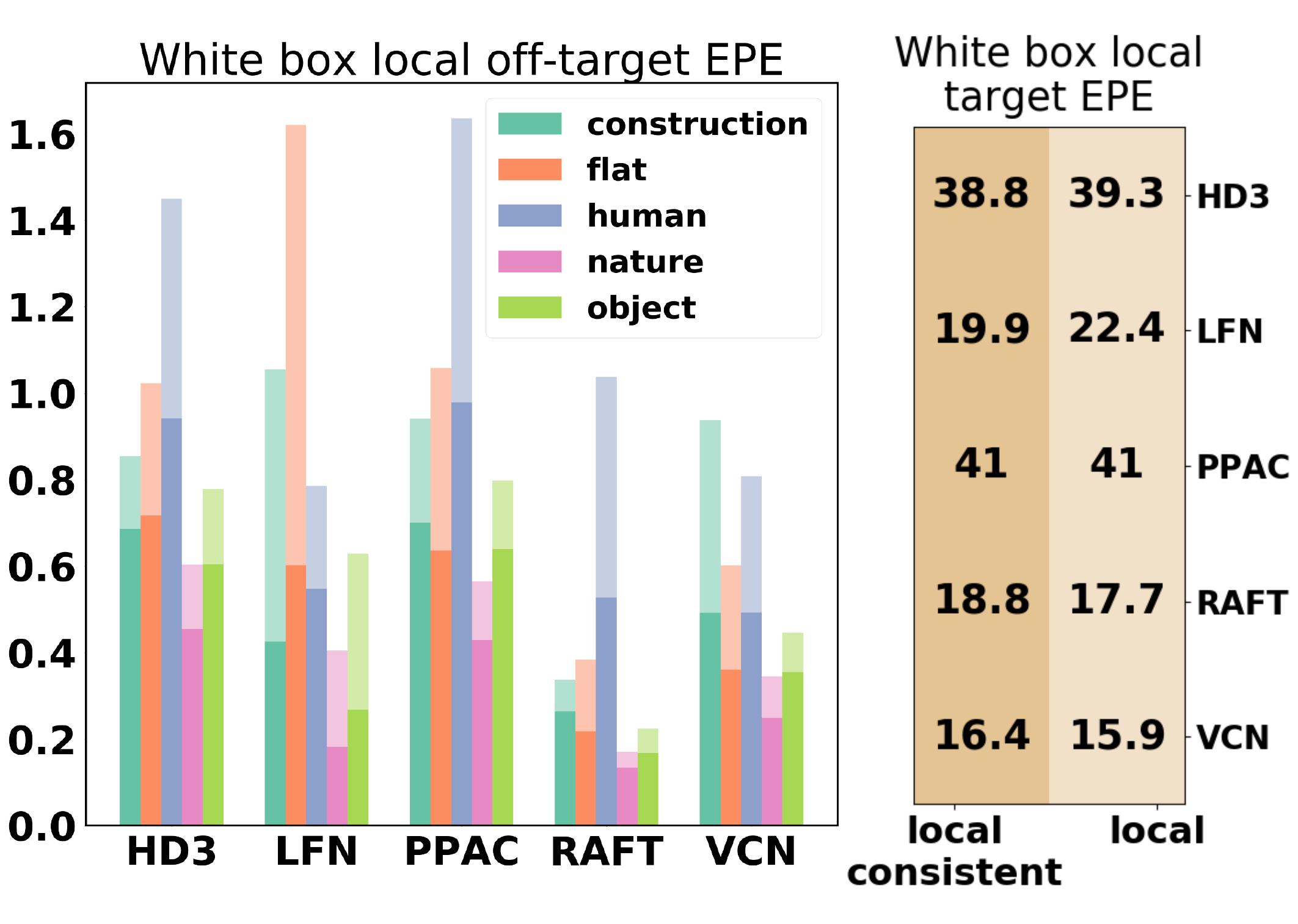} 
	\vspace{-0.8cm}
\end{center}
   \caption{Comparison between local attacks and local-consistent attacks on the KITTI dataset with $||\Delta{I}||=4 \cdot10^{-3}$. For each model, we attacked vehicle pixels and evaluated the mean error caused by a local attack (transparent colors) and a consistent local attack (solid colors) over the corresponding category. Consistent attacks reduce off-target damage and keep a similar effect on the target.}
\label{fig:Figure2Local}
\vspace{-0.4cm}
\end{figure}

\begin{figure*}[!t]
\begin{center}
   \includegraphics[width=1.0\linewidth]{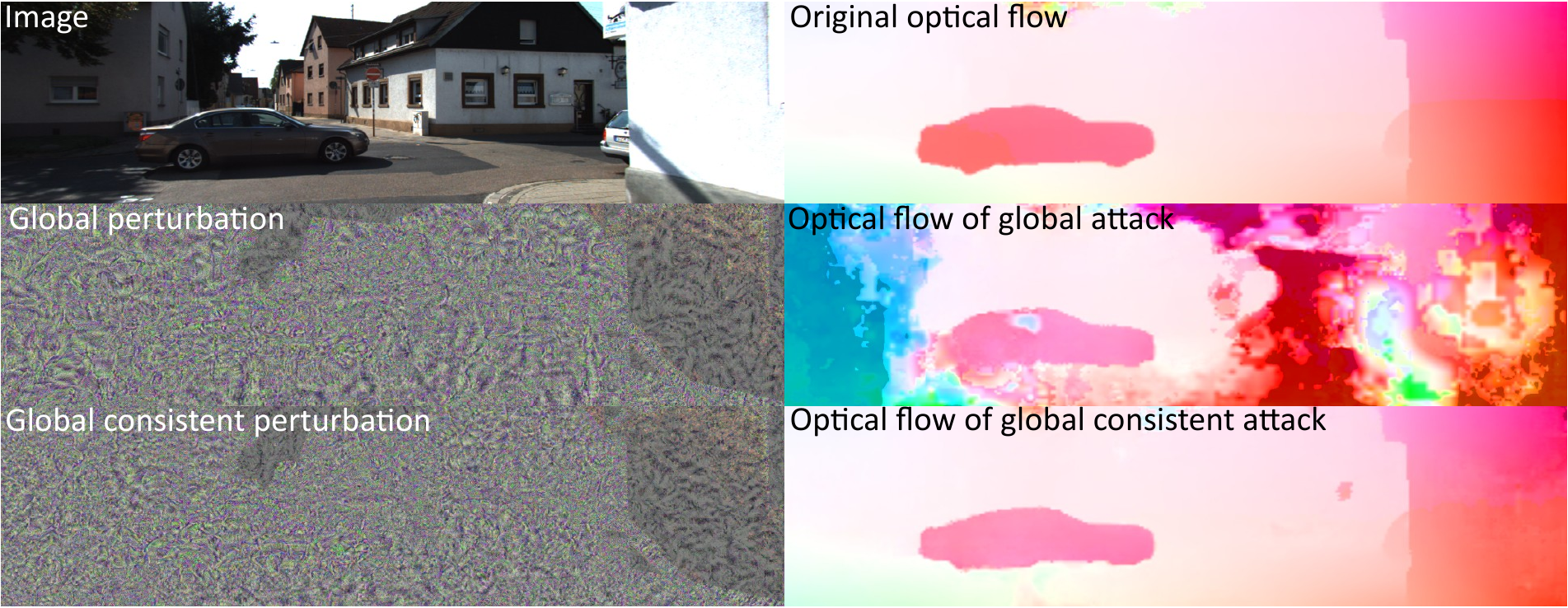} 
   \vspace{-0.5cm}
\end{center}
   \caption{A visualization of a global vehicle-targeting attack and the effect of adding a consistency term on HD3's \cite{HD3_19} flow with $||\Delta{I}||=2 \cdot10^{-2}$. Left: original image and perturbations. Right: the corresponding optical flows. Adding a consistency term reduces the effect on non-vehicle pixels while still significantly changing the vehicle optical flow}
\label{fig:Figure3Global}
\vspace{-0.5cm}
\end{figure*}

\begin{figure}[t]
\begin{center}
	\includegraphics[width=1.0\linewidth]{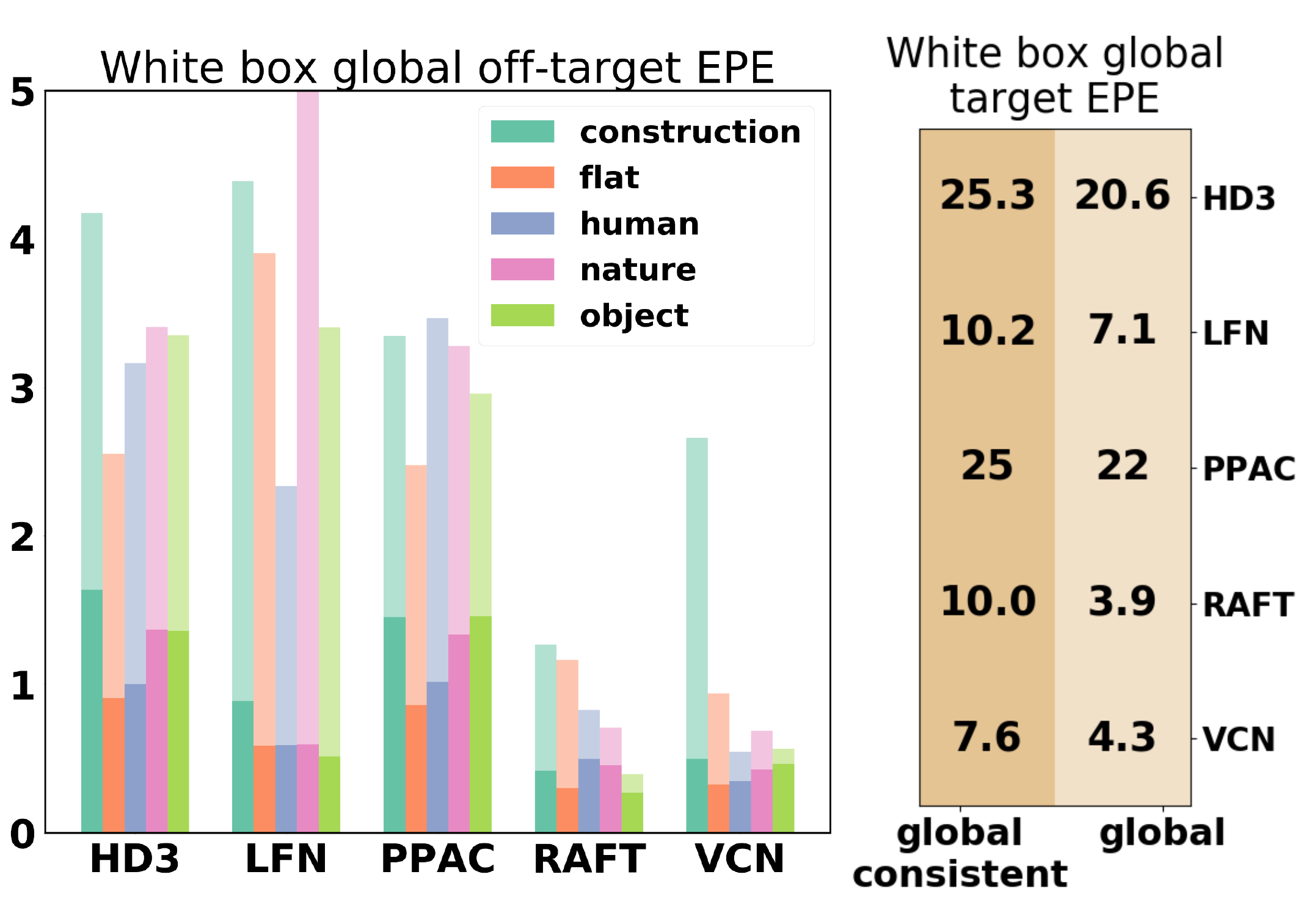} 
	\vspace{-0.8cm}
\end{center}
   \caption{Comparison between global attacks and global-consistent attacks on vehicles in the KITTI dataset with $||\Delta{I}||=4 \cdot10^{-3}$. For each model, we attacked vehicle pixels and evaluated the mean error caused by the global attacks (transparent colors) and the consistent global attacks (solid colors) over the corresponding category. Consistent attacks reduce off-target damage and significantly increase the effect on the target.} 
\label{tab:Figure4Global}
\vspace{-0.2cm}
\end{figure}

\begin{figure*}[!t]
\begin{center}
   \includegraphics[width=1.0\linewidth]{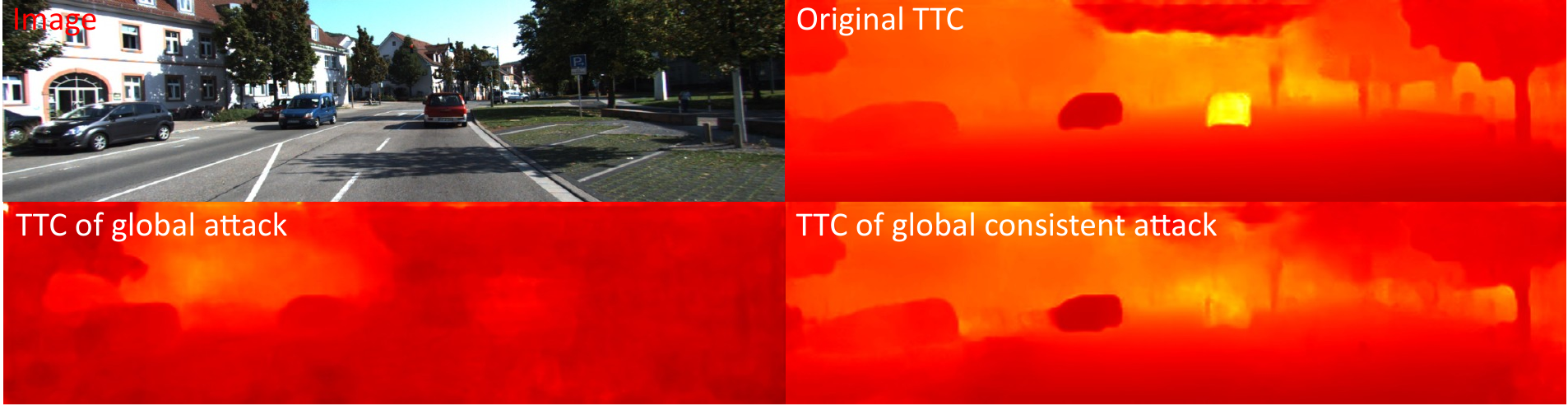} 
   \vspace{-0.7cm}
\end{center}
   \caption{Visualization of the TTC results for an AA on a vehicle instance using HD3 with $||\Delta{I}||= 4\cdot 10^{-3}$. Hot colors corresponds to shorter TTC than cold colors.}
\label{fig:quali_ttc}
\vspace{-0.2cm}
\end{figure*}

\begin{figure*}[th!]
\begin{center}

\includegraphics[width=0.49\linewidth]{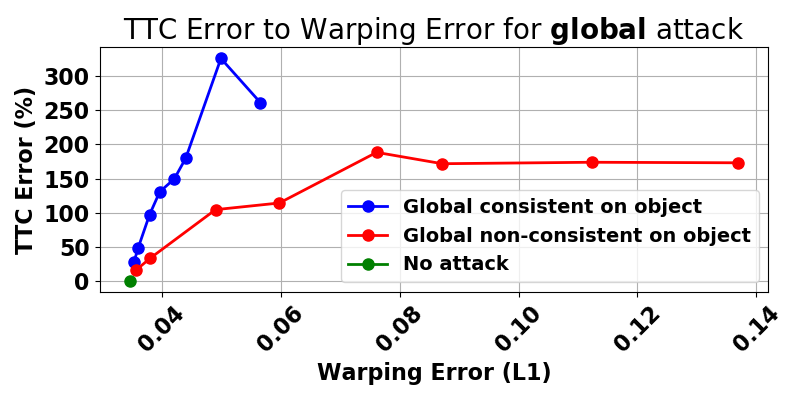}
\includegraphics[width=0.49\linewidth]{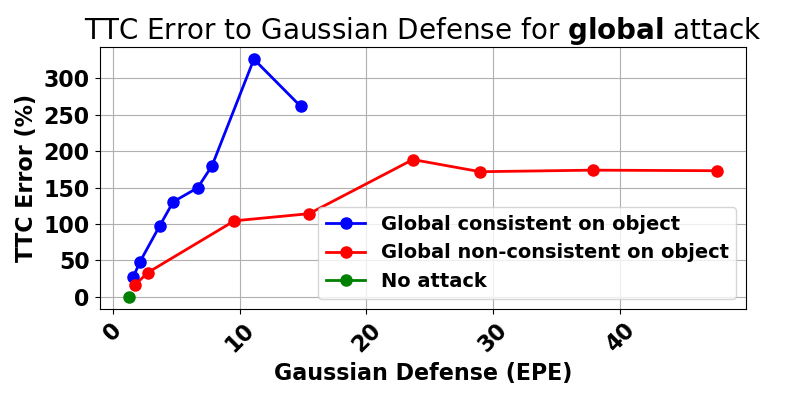} \\
\vspace{-0.1cm}
\includegraphics[width=0.49\linewidth]{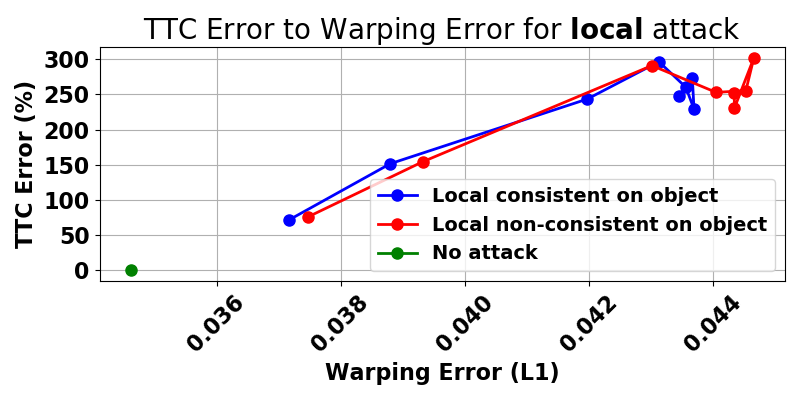}
\includegraphics[width=0.49\linewidth]{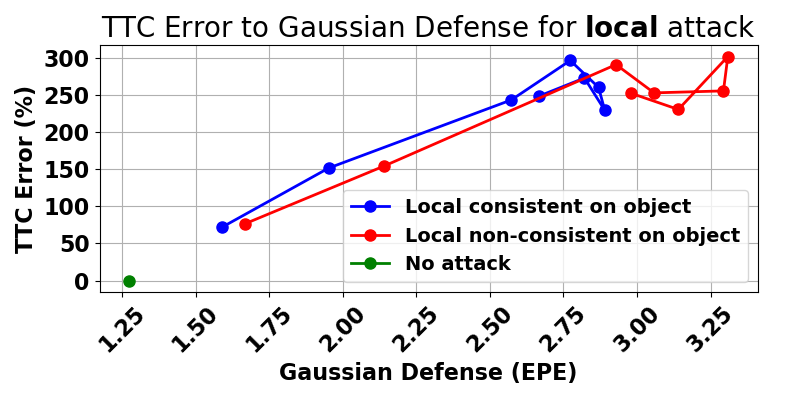}
\end{center}
\vspace{-0.5cm}
\caption{TTC error to AA detection score. The top and bottom rows correspond to the global and local attacks, respectively. The Y-axis in all graphs corresponds to the TTC error, while the X-axis corresponds to the AA detection score using warping error and Gaussian defense for the left and right graphs, respectively. The graph is created using attacks with magnitude $||m\cdot 10^{-3}||$ for $m \in \{0.2, 0.4, 1.2, 2, 3.2, 4, 6, 8\}$.}
\label{fig:ttc_to_detection}
\vspace{-0.2cm}
\end{figure*}

\begin{equation}
\begin{split}
& I_1^{(i)}=I_1^{(i-1)} + \delta I_1^{(i)} \\ 
& \delta I_1^{(i)}=\epsilon \cdot M_{perturb} \cdot \sign \left( \nabla l_{total}(V^{(i-1)},V)\right) \\
& V^{(i)}=f_{flow} (I_1^{(i)}, I_2)
\end{split}
\label{eq:eq4}
\end{equation}

In each attack step $i$ we create a small perturbation to the first image $\delta I_1^{(i)}$. As shown in Figure \ref{fig:FigureMethod} this perturbation is only applied in pixels $(x,y)$ where $M_{perturb}(x,y)=1$. It is equal to the sign of the loss function's gradient, weighted by the attack coefficient $\epsilon$. After computing the $i$'th perturbation we add it to the image from the previous step $I_1^{(i-1)}$ to get the current perturbed input $I_1^{(i)}$. Inferring on this input with the optical flow network $f_{flow}$ results in the $i$'th attacked optical flow map $V^{(i)}$. The loss between this flow $V^{(i)}$ and the original flow $V$ will then be used to compute the perturbation for the following step.
It is worth to note that for the first iteration ($i=1$) we add a small amount of white noise to the original flow $V$ so we would have non-zero gradients.

Let us define the target L1 norm of the perturbation as  $||\Delta{I}||$. Given the number of perturbed pixels $L$ and an estimated number of steps $n$ for the attack, we set $\epsilon$ according to:
\begin{equation}
\epsilon = \frac{||\Delta{I}||} {n \cdot L}
\label{eq:eq5}
\end{equation}
We then iteratively update our input using Equation \ref{eq:eq4} until $||I_1^{(i)}-I_1||_1 \approx ||\Delta{I|}|$ (up to 5\%).
We use $n=2$ and $||\Delta{I}||=4 \cdot10^{-3}(\approx 1/255)$ for most of our experiments, and will specifically state experiments with other values.

\subsection{Implementation details}

Throughout our experiments we use five optical flow models to evaluate the impact of adding our consistency term on targeted category-specific adversarial attacks – HD3 \cite{HD3_19}, PPAC \cite{PPAC_20}, VCN \cite{VCN_19}, RAFT \cite{RAFT_20}, LFN \cite{LFN_18}.
These models are some of the top performing methods on the KITTI''15 \cite{KITTI_15} dataset.
We use the published, pre-trained models, given by the authors of each of the five chosen models.
Since some models published multiple checkpoints, we always use the one fine-tuned on KITTI for our attack.

All of our experiments are performed and evaluated on the KITTI 12' \cite{KITTI_12} and KITTI 15' \cite{KITTI_15}  datasets. These datasets contain a semantic segmentation labeling that we employ in our attacks.
We could have used any semantic segmentation method \cite{semseg_chen2017rethinking, semseg_chen2017deeplab, semseg_chen2018encoder} to label each scene. This would simulate a more realistic scenario where ground truth labeling is unavailable.

We evaluate our attack results using the average \textit{end-point-error} (EPE) metric \cite{Middlebury_BakerSLRBS11}, which computes the average $L2$ norm of the difference between attacked and original flows. 
The averaging is usually done over all image pixels, but since we are particularly interested in the effect of our attack on semantic classes, we compute the EPE averaged on pixels of specific classes.
Using this metric we can estimate the average shift in OF prediction due to the attack, for each class of interest.

In the subsequent section we will elaborate on the results from our main experiments. 
These experiments will encapsulate three different attack settings. These settings differ in the perturbed pixels mask ($M_{perturb}$, defined in Equation \ref{eq:eq4}) and the pixels we aim to attack ($M_{target}$, defined in Equation \ref{eq:eq1}). In the first setting, a local attack, we perturb vehicle pixels and aim to attack the same subset of pixels. The second setting, a global attack, is where we perturb the entire image, but aim to attack vehicle pixels only. The third setting, a cross-category attack, is where we perturb the pixels of nature pixels, and aim to attack vehicle category pixels (presented in the Supplementary material).


\section{Experiments and Results}
In this section, we present the experimental results obtained for the "vehicle" target category. The results for "human" target category, as well as the results obtained using the KITTI 12', are given in the supplementary.
\subsection{Local attacks}
Figure \ref{fig:Figure1Local} visualizes an example local attack ($||\Delta{I}||=2 \cdot10^{-2}$) experiment using the LFN model \cite{LFN_18}. In this experiment a vehicle instance was attacked by only perturbing its pixels. Two attacks were conducted: a baseline, non
-consistent, method with $\alpha=0$ and a consistent attack with $\alpha=10$.

Both attacks are successful in manipulating the car's targeted flow and cause the previous right (red) moving vehicle to turn left (blue). However, the consistent attack preserves the non-targeted flow better, as can be seen by comparing the flow under the vehicle. 

To quantify this effect, this experiment was expended to the entirety of the KITTI dataset. Here, for each image in the dataset we have attacked all of the vehicles in that image (by perturbing vehicle pixels). We then evaluated the mean EPE between original and attacked flow on selected categories: construction, flat, human, nature, object and vehicle.

Figure \ref{fig:Figure2Local} presents the results on the KITTI dataset using the five selected models. 
The left sub-figure presents the (undesired) effect on non-targeted pixels and the right sub-figure presents the (desired) effect on targeted pixels.
We see that while the targeted vehicle category error does not vary much between attacks (right table) the non-targeted categories (left figure) suffer much less damage using a local consistent attack than our baseline non-consistent attack.
The right sub-figure, that presents the targeted EPE, shows a small difference between attacks.
The left sub-figure, however, shows a much larger difference. 
The effect on non-targeted categories is significantly reduced using our consistent attacks. In particular there is a 35\% decrease on average (across methods) on the error induced on these categories.

\subsection{Global attacks}

One of the concerns with using a local attack is that since we perturb only a subset of the image pixels, 
we employ a high $L_{\infty}$ norm to achieve the same $L_1$  norm as a \emph{global} attack that perturbs the entire image. This, in turn, causes the local attack to be more perceptible compared to a 
global attack. 
Figure \ref{fig:Figure3Global} demonstrates this global attack in which we perturbed the entire image. The figure visually compares the results of the consistent and non-consistent attacks.
The left column shows the original image and its perturbations. Here, unlike the local attack, the entire image is perturbed. The right column shows the effect both attacks have on HD3's optical flow. For the non-consistent attack we can notice multiple non-vehicle flow segments that changed drastically, turning the naturally smooth flow of the background into a rapidly varying flow.
Repeating the methodology we used for the local settings, we expand this experiment by attacking all of the vehicle category in the KITTI dataset \cite{KITTI_15}, and averaging the error over the selected classes. 

Figure \ref{tab:Figure4Global} shows the result of attacking all vehicles in a global setting over the KITTI dataset, for our five OF models with $||\Delta{I}||=4 \cdot10^{-3}$. Similarly to the local case, the left and right sub-figures demonstrate the effect on the non-targeted pixels and targeted pixels, respectively. The resulting targeted vehicle category error is higher when using a consistent attack (right table). Moreover, the non-targeted categories (left figure) suffer significantly less damage using a global consistent attack than the baseline non-consistent attack. Thus, for example, using the consistent attack results in a 60\% stronger effect on the targeted category (averaged across models), while removing 60\% of the unwanted optical flow change on the remaining categories (averaged across models).


\subsection{Time-To-Collision (TTC)}
As discussed in Section~\ref{sec:intro}, we emphasize the significance of adversarial attacks on OF models by their possible impact on TTC algorithms~\cite{yang2020upgrading, pedro2021collision, blumenkamp2019end, badki2021binary}. For this experiment, we chose the state-of-the-art TTC algorithm presented in \cite{yang2020upgrading}, which uses OF to compute a per-pixel TTC. We supply the model our attacked OF instead of its original OF predictions computed by the pre-trained VCN (without fine-tuning).

An attack on a vehicle instance, which is visualized in Figure~\ref{fig:quali_ttc}, demonstrates the impact of the original flow, the global consistent and global non-consistent attacked flows, on the TTC.
The TTC values are log-scaled and color-coded, where hot colors (redish) encode lower TTC than colder colors (yellowish-whitish).
The attacked vehicle, which is yellow (high TTC) in the original flow, is significantly darker (low TTC) in both the consistent and non-consistent attacks.
Importantly, the backgrounds of the original and consistent attack are quite similar, while the background of the not-consistent attack is very different.

As argued in Section~\ref{sec:intro}, the effect of the off-target consistency loss term allows a better tradeoff between the impact on the TTC and a AA detection score. 
An example for this tradeoff would be that an attacked input with the same detection score will result in a higher average TTC impact. 
To quantify this tradeoff we've used three AA detection methods.


\vspace{-0.4cm}
\paragraph{Warping error:} The difference between $I_1$ and $I_2$ warped using the (attacked) OF $V$. That is, $||W_V(I_2)-I_1||_1$, where $W_V(I_2)$ warps $I_2$ using the OF~$V$. The warping error is often used as an OF confidence measure  \cite{Learningmotion}. Naturally, such confidence measure may be used to estimate an AA detection score.   
\vspace{-0.4cm}
\paragraph{The Gaussian/Median defenses \cite{xu2017feature}:} The OF error (EPE) between the predicted flow and the flow from Gaussian/Median smoothed versions of the same images. That is, let $V'(I_1',I_2)$ be an attacked flow, and $V_K'=V'(K(I_1'),I_2)$ be an attacked flow (using the same attack) with $I_1'$ smoothed using a $3\times 3$ Gaussian/Median kernel $K$ before the OF computation. $||V'-V_K'||_1$ is used to estimate the detection score. Such defense methods were used as AA detection methods in \cite{xu2017feature}.


The graphs for the error in TTC as a function of the AA detection score are presented in Figure~\ref{fig:ttc_to_detection}. (The median defense is presented in the supplementary material.)
We measure the error in TTC as an average percentage of difference relative to the original TTC; that is, $|T_A-T_O|/T_O$, where $T_A$ and $T_O$ are the TTCs of the attacked and original flows, respectively. 
The graphs are created from 8 AA with different magnitudes, where all 5 OF models (in a white-box settings) are averaged per attack magnitude.
In all three cases, the global consistent attack is superior to the global non-consistent attack in both the detection score (lower in X axis), and in impact on TTC (higher Y axis).
In the local attack, the trend is similar, however, the gap is much smaller. 
To conclude, the off-target consistency loss term is effective in terms of the TTC - detection score tradeoff.  


\begin{figure}[t]
\begin{center}
   \includegraphics[width=1.0\linewidth]{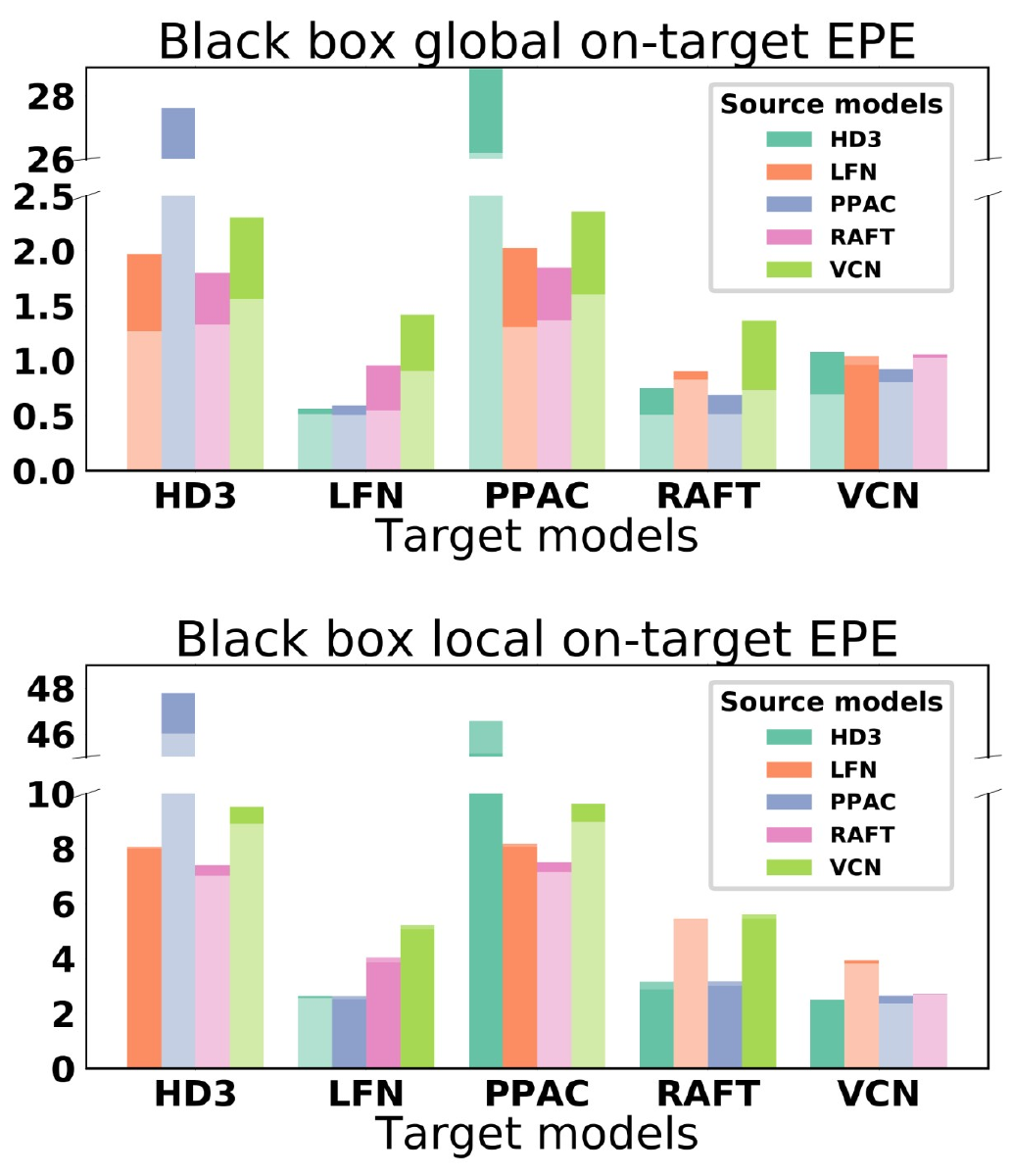}
\vspace{-0.8cm}
\end{center}
   \caption{ "Black-box" setting results for the global attacks (top) and local attacks (bottom) on vehicles with $||\Delta{I}||=4 \cdot10^{-3}$. Bar height signify the mean EPE over the vehicle category caused by an attack created using the source models (color-coded) on the target model (x-axis). Consistent attacks are marked by solid colors and unconstrained attacks by transparent colors.  Consistent attacks significantly increase transferability for the global setting. For the local setting only small variations are measured. }
\label{fig:FigureBlackBox}
\vspace{-0.3cm}
\end{figure}

\subsection{Black-box attacks and transferability}

Finally, we evaluated the transferability of the consistent attacks for the global and local attacks. To this end, we used each of the chosen models to attack the vehicle category in every image pair in the KITTI dataset.
This resulted in 5 adversarial KITTI datasets created using our baseline attack, and 5 datasets created using our consistent attacks. 
Each model was then evaluated on the adversarial datasets generated using the remaining models. The mean EPE over the vehicle targeted category for each attack is presented in Figure \ref{fig:FigureBlackBox}. We use transparent colors to visualize non-consistent attacks and solid colors to visualize for consistent attacks. 
We note that attacks created using HD3 seem to have a high impact on PPAC and vice versa. This could be related to HD3 and PPAC having most of their architecture shared. They are too similar to be considered a “black-box” attack in the classical sense, and hence they greatly impact one another.

Similar to the results we got in the white box settings, the local attack impact does not vary a lot with the addition of the consistency term.
However, for the global case we observe a significant increase in the targeted impact transferred to other models. 
If we examine the results on RAFT, adding the consistency term resulted in a 44\% increase in black-box attack strength, averaged across targeted models.


\section{Discussion}

To summarize, we presented a new methodology for targeted adversarial attacks against optical flow models. We introduced a new term to the attack, called “consistency term”, which is used to reduce the effect of the attack on the off-target pixels. In three different settings: local, global and cross category (supplementary), adding the consistency term to the loss reduces the impact on non-targeted object. Adding the term either preserves or increases the effect on the targeted category (depending on the setting). Moreover, we have demonstrated that for some of the settings using a consistent attack results in a more transferrable attack. Finally, we have showed that for a TTC downstream task these attacks have a better detection - impact tradeoff, with an impact as high as three times higher for the same detection score.

In our experiments we observe an obvious difference between the local and the global setting.
 First, the effect on the non-targeted object is much more apparent in the global setting.
Indeed, in this setting, the danger of negatively impacting the rest of the scene is much greater since we directly change the non-target pixels.
Adding the consistency term allows us to introduce global perturbations with a smaller effect on the resulting non-targeted optical flow.
 
An interesting follow-up for our work would be utilizing adversarial targeted attacks as a data augmentation technique for model training. Recent optical flow models have demonstrated the benefits of data augmentation in the training procedure \cite{dataaugment_liu2020learning}. Other works \cite{wong2020targeted} have demonstrated that some semantic classes are easier to attack than others. By leveraging consistent adversarial targeted attacks in its augmentation procedure, models might be able to learn a more robust representation of each semantic class. This, in turn, might decrease the probability of a successful attack against them \cite{understood_madry2017towards}, and increase the ability of a model to generalize its predictions for those classes \cite{stutz2019disentangling}. 

\newpage

\section{Supplementary Material}
In this supplementary we provide additional material on consisted targeted attacks. First, we provide the details of the cross category attacks against the vehicle category conducted on KITTI 15'. Then, we analyze the effect of varying $\alpha$ on attack EPE. Following that, we provide our results on both the human category attack on KITTI 15' and the vehicle category attack on KITTI 12'. The results for the TTC as a function of the Median AA detection score is then presented. Finally, we provide further attack visualizations for the experiments conducted on KITTI 15'

\subsection{Cross category attacks}
\begin{figure}[t]
\begin{center}
   \includegraphics[width=0.99\linewidth]{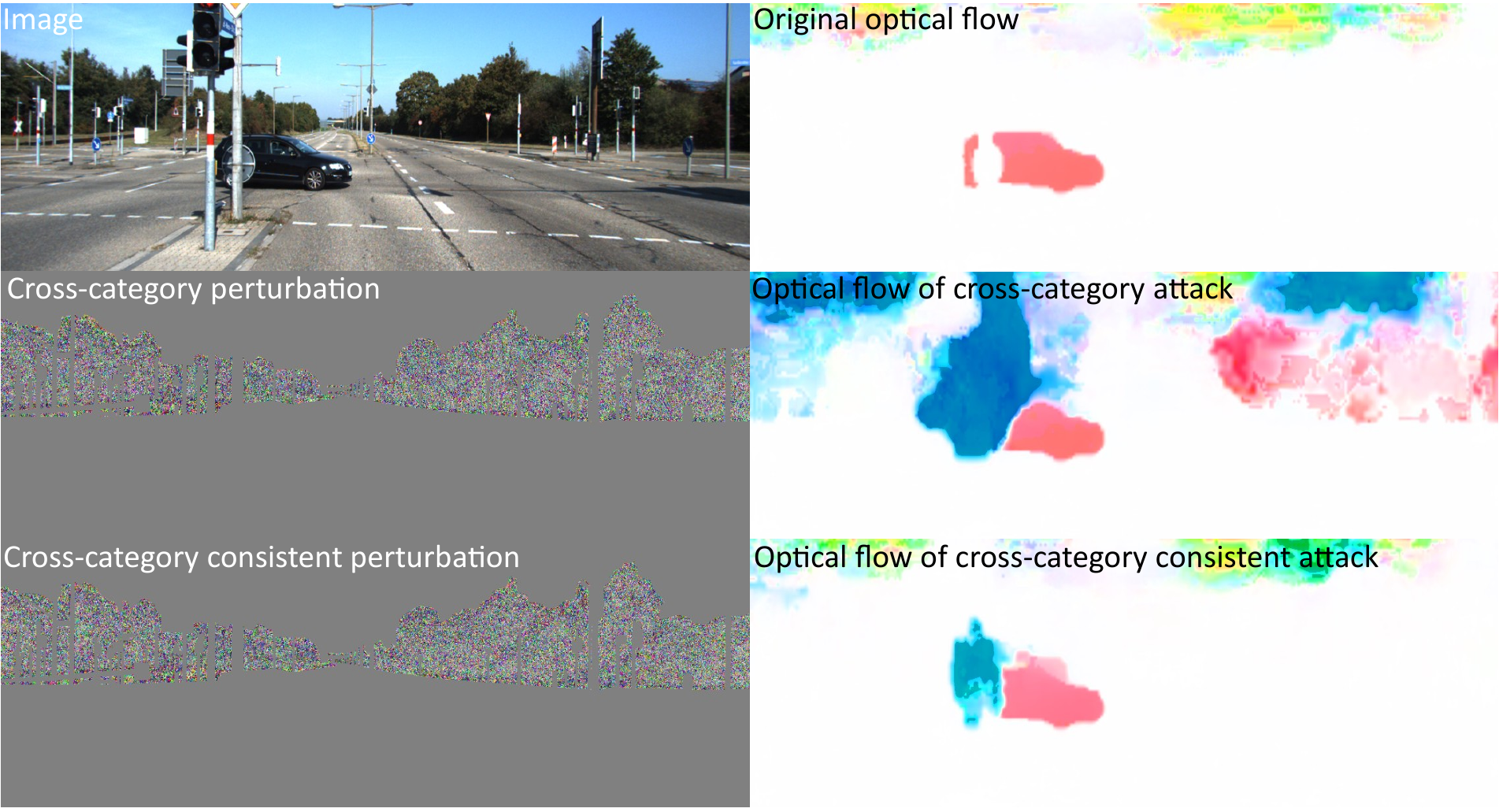}
\end{center}
   \caption{ A visualization of a cross-category attack baseline and the effect of adding a consistency term on a vehicle instance using HD3 with $|\Delta{I}|= 6.5 \cdot10^{-2}$. While the original flow of the vehicle indicate a right-moving instance, the attacked flows also indicates a left moving instance. In the non-consistent case (second row) we see that the remaining scene is affected as well, and once a consistency term is being added (last row) the attack is much more focused on the attacked instance }
\label{fig:Figure4CrossClass}
\end{figure}

In the cross-category setting, we perturb the nature category pixels in order to manipulate the optical flow of vehicle pixels. Figure \ref{fig:Figure4CrossClass} visualizes such an attack and shows the original input and flow, compared to the cross category and consistent cross category attacks. 
First, the second row of Figure \ref{fig:Figure4CrossClass} shows that our non-consistent attack results in a significant change to both the car’s flow and its environment. Both the sky, the nature pixels, and large portions of the road are affected by this attack. Then, once consistency is added (third row), the changes to the non-vehicle pixels are much less apparent, with the vehicle’s flow still significantly changed.

\begin{figure}[th]
\begin{center}
   \includegraphics[width=0.99\linewidth]{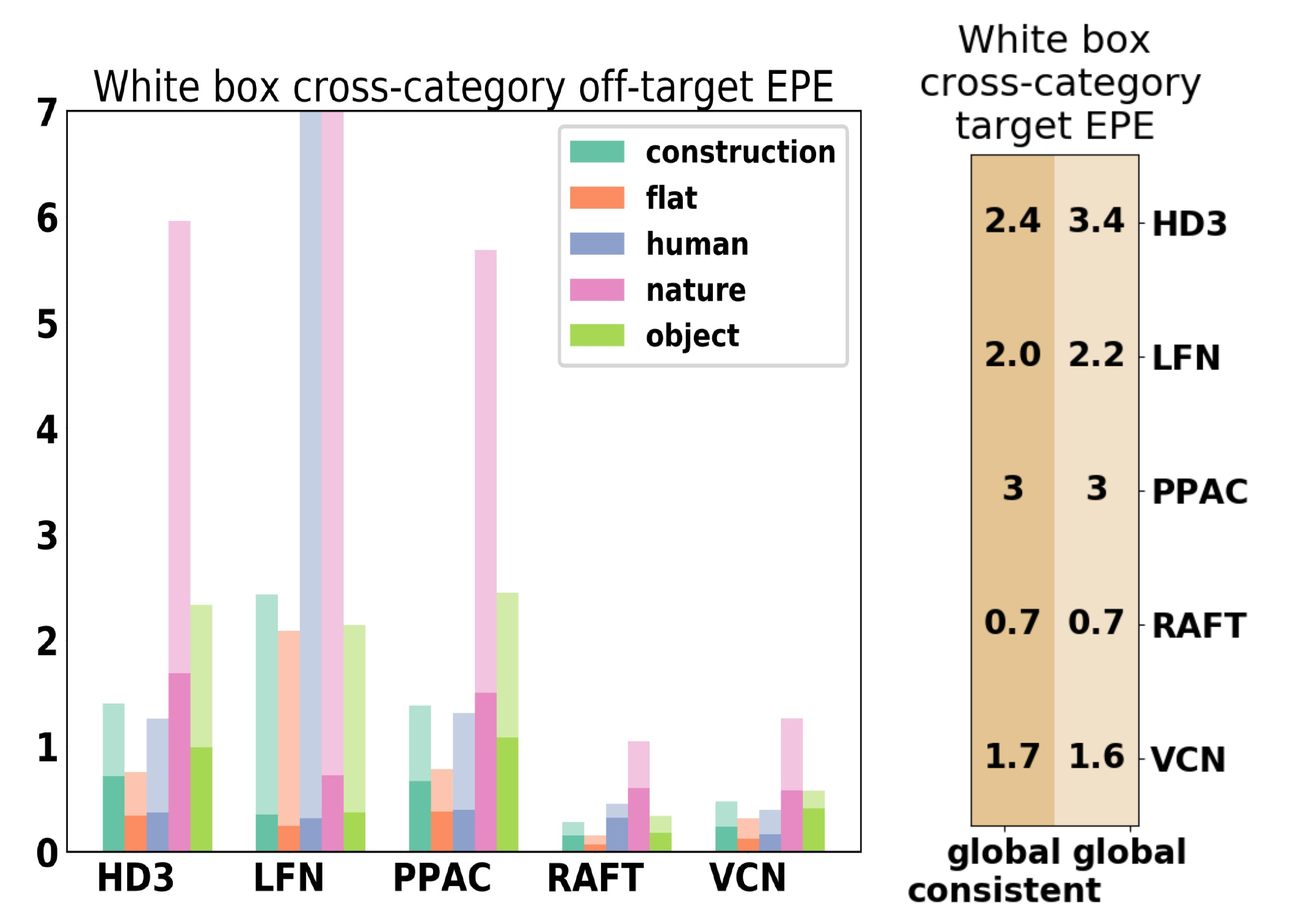}
\end{center}
   \caption{ Comparison between cross category attacks and consistent
cross category on vehicles in the KITTI dataset with $|\Delta I|=4 \cdot 10^{-3}$. For each model, we attacked vehicle pixels by perturbing nature pixels. We then evaluated
the mean error caused by the non-consistent attack (clear colors) and
the consistent attack (solid colors) over the corresponding
category. Consistent attacks reduce off-target damage and keep attack efficiency similar. The effect on off-target is especially high on the perturbed nature category.}
\label{fig:cross_Category}
\end{figure}

We've evaluated the cross category effect on the KITTI 15' dataset. Here, we've used $|\Delta{I}|= 4 \cdot 10^{-3}$ to perturb nature pixels. The target of the attack was vehicle pixels. Figure \ref{fig:cross_Category} presents the evaluation of this experiment. It shows the EPE comparing to the original flow on the off-target categories (left) and the on-target vehicle category (right).
 Two distinct feature of this attack settings are noticeable in the left figure. First, all of the categories that were neither attacked nor perturbed (human, flat, construction, object) are less effected from the consistent attack. Second, the perturbed category (nature) is highly effected in the non-consistent attack. This, since nature pixels are altered in this attack. Adding a consistency term greatly reduce the error we observe on the perturbed category.

\subsection{ Varying the consistency parameter }

In the method section we've introduced the consistency parameter, $\alpha$. This parameter controls a trade-off of the consistent attack. High values of $\alpha$ lead to an attack that is focused on preserving non-target flow. Low values focus instead on damaging the optical flow of the target flow.

To demonstrate the trade-off $\alpha$ controls and the role of the consistency term as a regularization term we've conducted the following experiment. We've repeated the same global attack against vehicles described in our work with various values of $\alpha$. These values range from $\alpha=0.01$ to $\alpha=100$. For each attack we've measured two metrics. First, the mean EPE (compared to original flow) on all target pixels. Second, the mean EPE on all non-targeted categories: construction, flat, nature, object, human. 

\begin{figure}[th]
\begin{center}
   \includegraphics[width=0.99\linewidth]{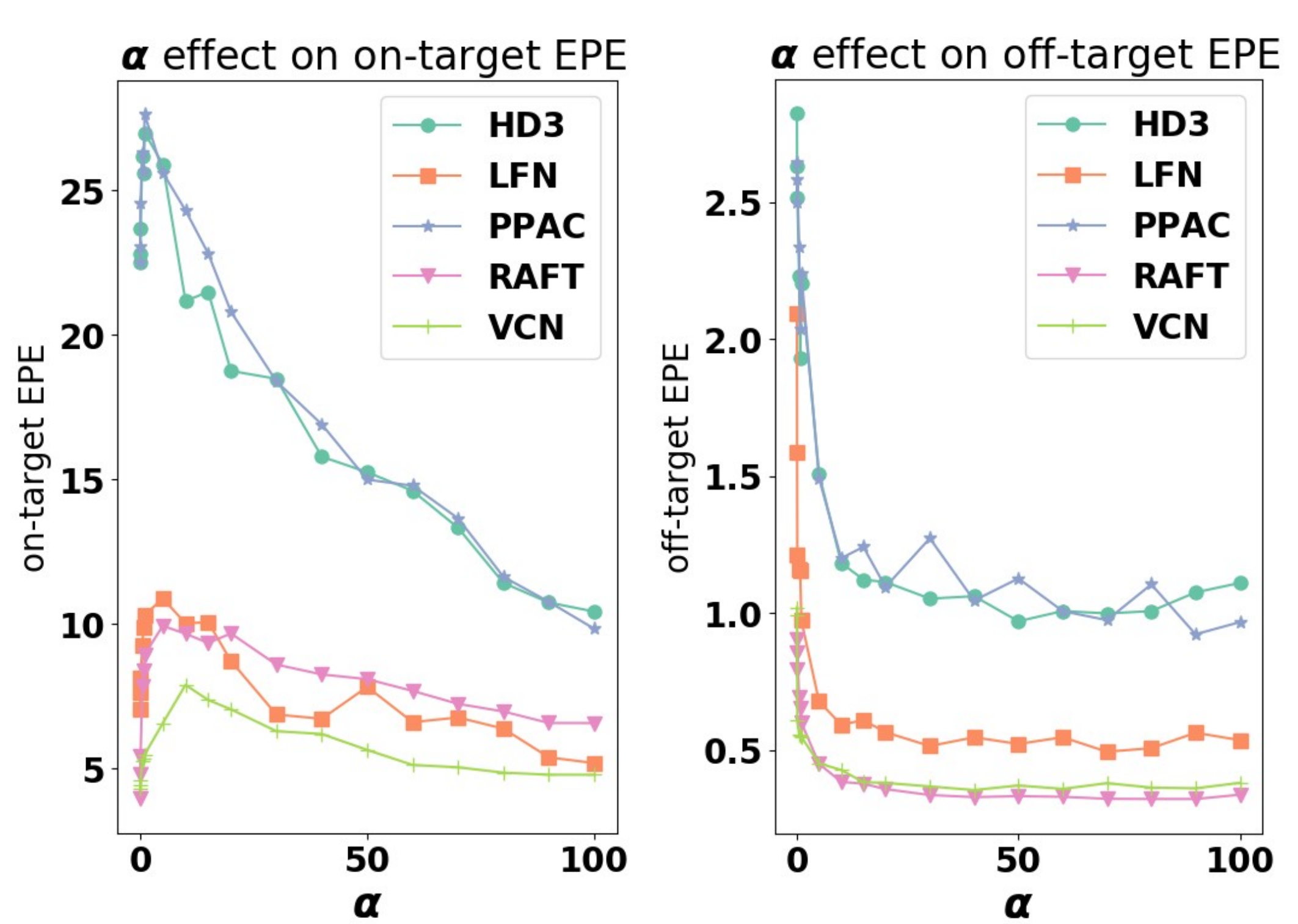}
\end{center}
   \caption{ The effect of $\alpha$ on attack metrics. Varying $\alpha$ effects both on-target EPE (left) and off-target EPE (right). Left figure shows that increasing $\alpha$ initially increase attack efficiency and then reduces it. Right figure shows that increasing $\alpha$ decreases our effect on non-targeted categories.}
\label{fig:alpha_experiment}
\end{figure}

The results of this experiment are presented in Figure \ref{fig:alpha_experiment}. The left figure presents the effect on the off-target EPE. It shows that the more we increase $\alpha$ the less we modify the non-targeted flow. Once $\alpha$ is large enough its regularizing effect seem to stable and the off-target EPE plateaus. The right figure presents the effect on target EPE. For small enough values of $\alpha$ there is an increase in the attack efficiency on the target. This is a result of $\alpha$'s role as a regularization. For small values, it constrains the system to find a more efficient attack.  For larger values of $\alpha$ we see a decrease in attack efficiency. This happens when the regularization term becomes very dominant. Instead of finding a more efficient perturbation we now focus the attack on preserving off-target flow. 

 \subsection{ Human attacks }
 
  \begin{figure}[t]
\begin{center}
   \includegraphics[width=0.99\linewidth]{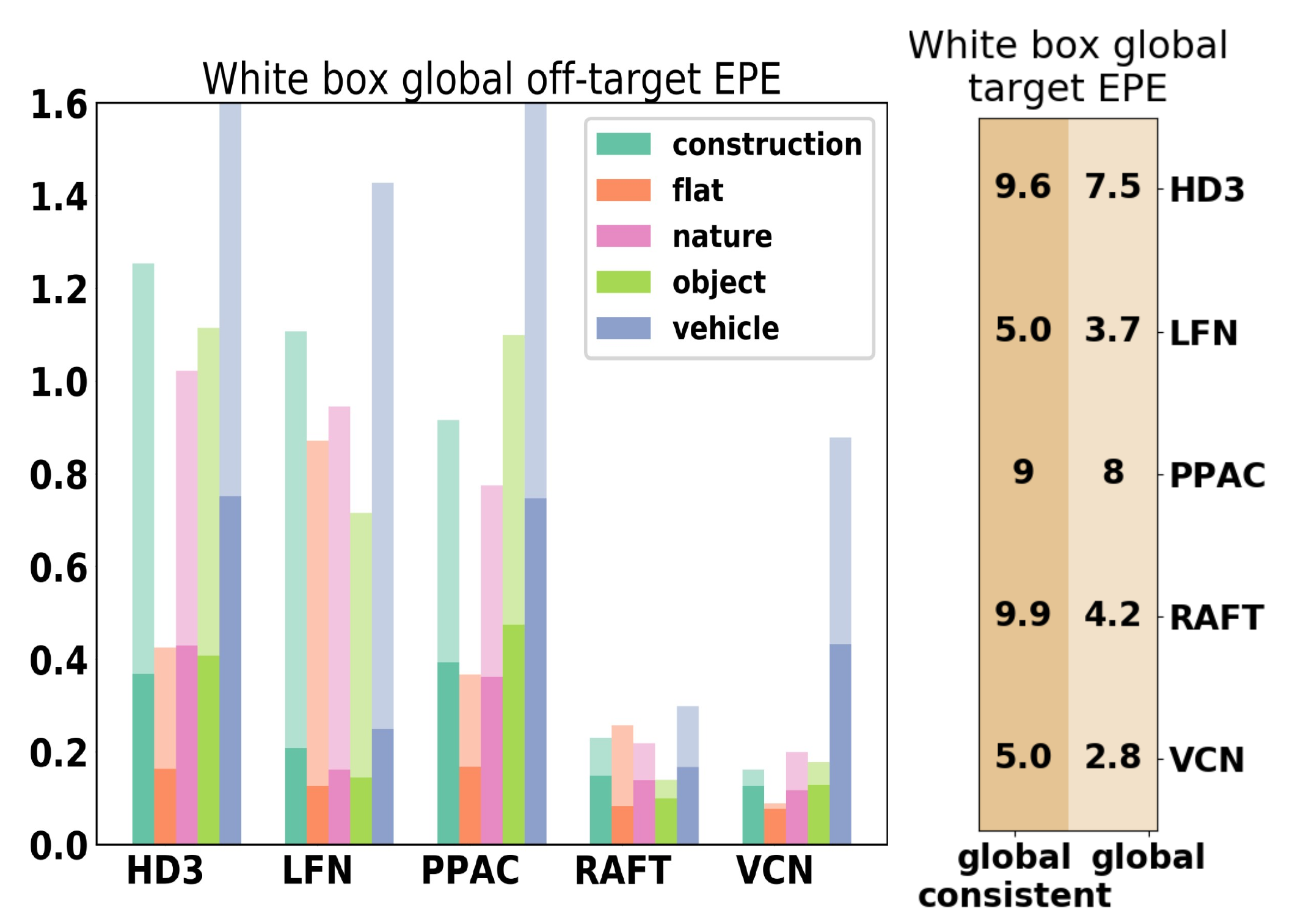}
\end{center}
   \caption{ Comparison between global attacks and global consistent
attacks on humans in the KITTI dataset with $|\Delta I|=4 \cdot 10^{-3}$. For each model, we attacked vehicle pixels by perturbing the entire image and evaluated
the mean error caused by the global attacks (clear colors) and
the consistent global attacks (solid colors) over the corresponding
category. Consistent attacks reduce off-target damage and keep attack efficiency similar. }
\label{fig:human_global}
\end{figure}

While we focused our experiments on the vehicle category, other categories could be attacked as well. Here, we evaluate our attack against the human category of the KITTI 15' dataset.
We use two settings, global and cross-category, to attack this category and report our results.
 We did not use the third, local, setting presented earlier. This is since humans are usually composed by a small amount of pixels. Thus, the amount of perturbation we can introduce when only perturbing human pixels is highly limited.
 
 We begin by reporting the global attack results on the human category. Here, we've perturbed the entire image in order to alter the optical flow of the human category pixels. Figure \ref{fig:human_global} presents the result of running this experiment on the entire KITTI 15' dataset. It shows the mean EPE (with respect to the original flow) averaged on the attacked, human, category (right). It also presents the same metric averaged on non-targeted categories (left).
 Here we see two effects previously demonstrated on the vehicle category. First, the consistent attack resulted in much less damage to the off-target categories. Thus for example, for HD3 the flat category error has decreased by ~50\%. Moreover, we see an increase in attack efficiency on the target. Thus, for example, using global consistent attacks increases the mean error on target pixels by 58\% on average.
 
\begin{figure}[t]
\begin{center}
   \includegraphics[width=0.99\linewidth]{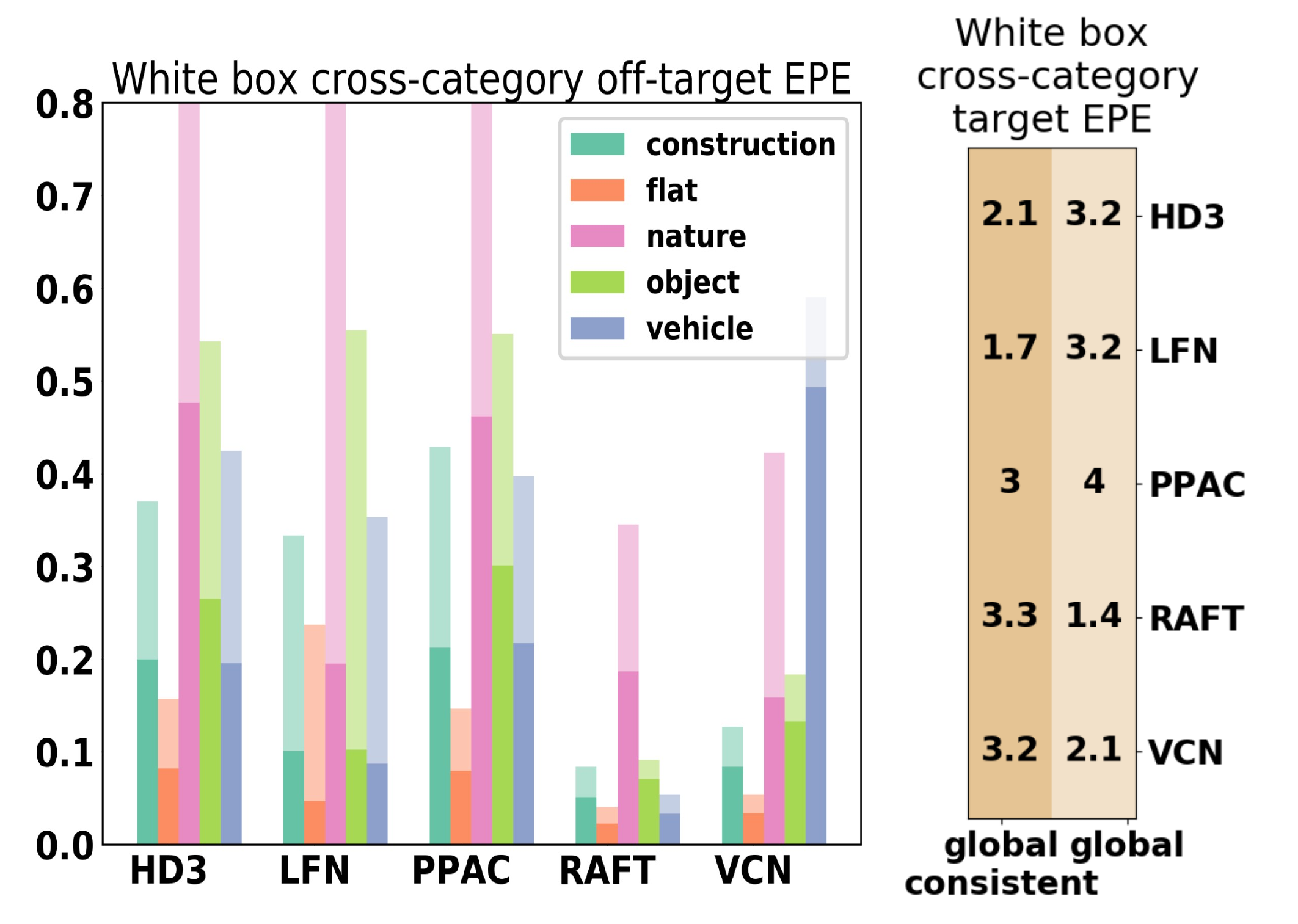}
\end{center}
   \caption{ Comparison between cross category attacks and consistent
cross category on humans in the KITTI dataset with $|\Delta I|=4 \cdot 10^{-3}$. For each model, we attacked humans pixels by perturbing nature pixels. We then evaluated
the mean error caused by the non-consistent attack (clear colors) and
the consistent attack (solid colors) over the corresponding
category. Consistent attacks reduce off-target damage and keep attack efficiency similar. The effect on off-target is especially high on the perturbed nature class.}
\label{fig:human_cross}
\end{figure}

 Next, we conducted a cross-category attack against the human category. In this setting we perturbed nature pixels to damage the optical flow of the human category. Figure \ref{fig:human_cross} presents the result of this experiment on the KITTI 15' dataset. First, we see that the off-target categories error reduce using this attack. Thus, for example LFN reduced its error on the vehicle category by approximately 75\%. The attack efficiency, however, remained similar for both attacks.

\subsection{ KITTI 12' dataset attacks }

Throughout our work we have presented multiple experiments conducted using the KITTI 15' dataset. Here, we utilize an additional dataset, KITTI 12' \cite{Geiger2012CVPR}, to evaluate consistent attacks against optical flow. The KITTI 12' optical flow dataset contains 193 image-pairs. Among those images, only the first 65 contain pixel-wise segmentation ground truth \cite{Geiger2012CVPR}. The rest of the images contain only vehicle labeling.

Since our experiments require pixel-wise segmentation, we restrict our evaluation on the KITTI 12' dataset to the first 65 images. We report our results on these images using the same evaluation protocol described in our method section. The consistent attack is evaluated under three settings: global, local, and cross-category attacks. 

 \begin{figure}[t]
\begin{center}
   \includegraphics[width=0.99\linewidth]{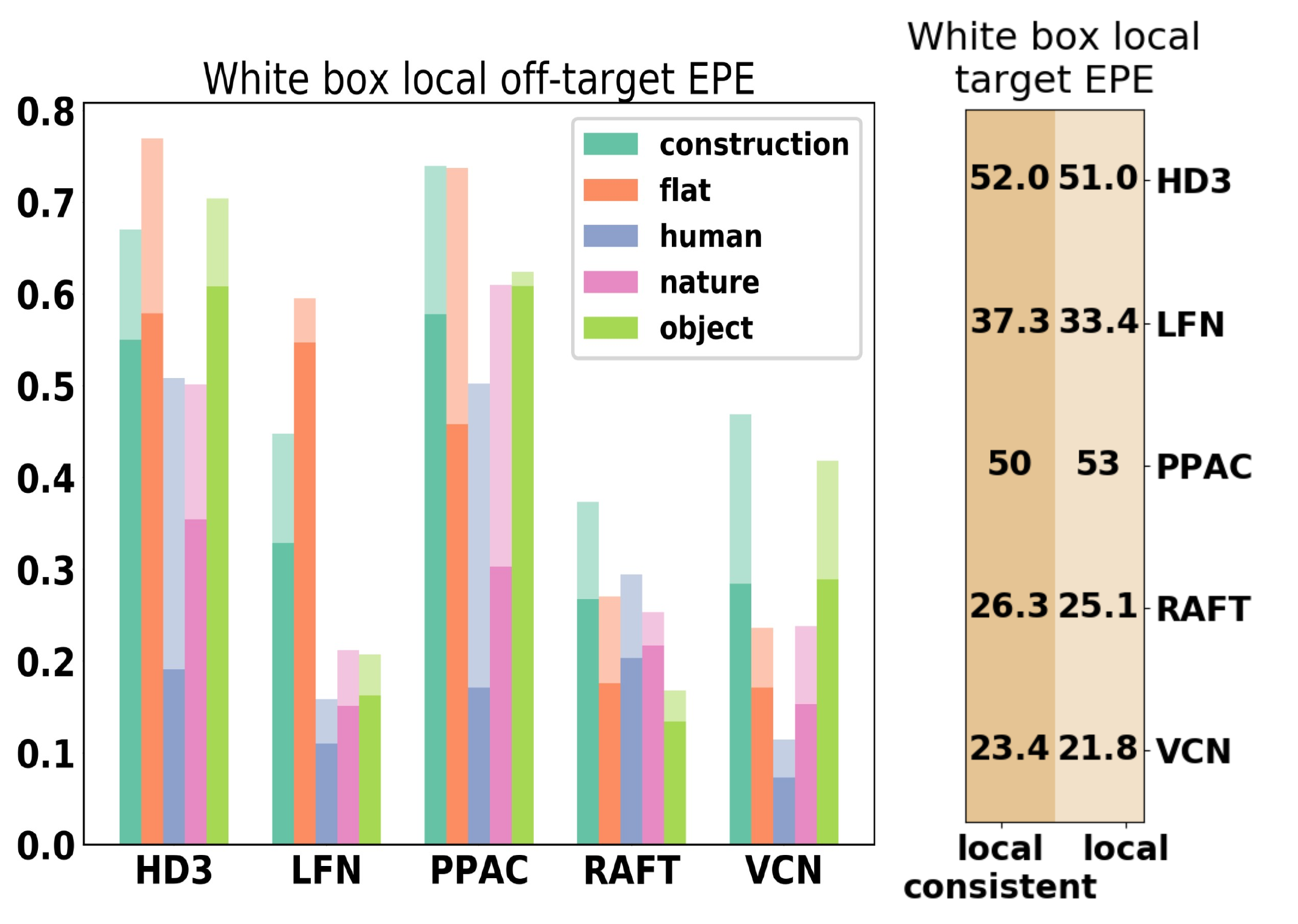}
\end{center}
   \caption{ Comparison between local attacks and local consistent
attacks on vehicles in the KITTI 12' dataset with $|\Delta I|=4 \cdot 10^{-3}$. For each model, we attacked vehicle pixels by perturbing vehicle pixels and evaluated
the mean error caused by the global attacks (clear colors) and
the consistent global attacks (solid colors) over the corresponding
category.}
\label{fig:KITTI12_local}
\end{figure}

 \begin{figure}[t]
\begin{center}
   \includegraphics[width=0.99\linewidth]{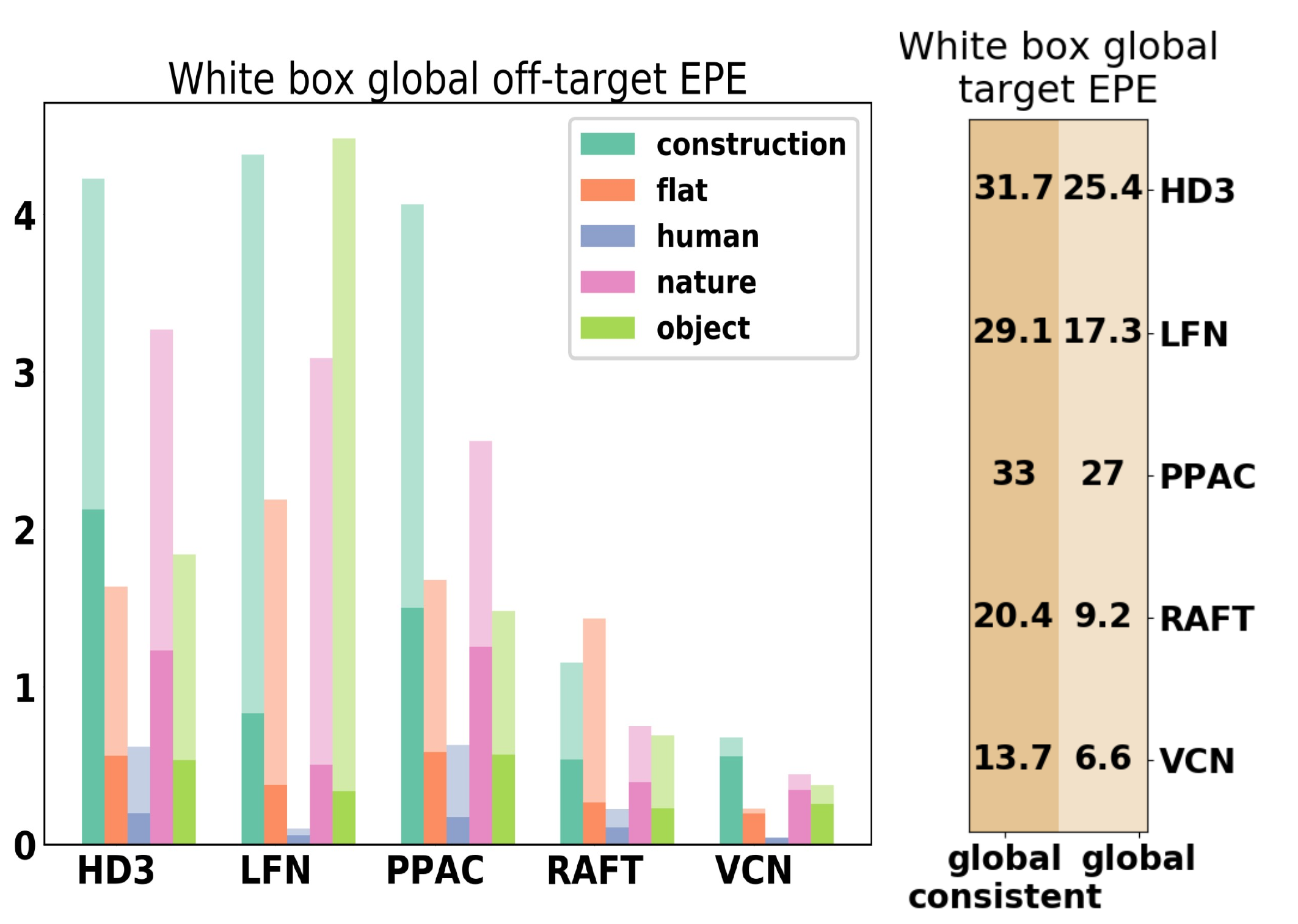}
\end{center}
   \caption{ Comparison between global attacks and global consistent
attacks on vehicles in the KITTI 12' dataset with $|\Delta I|=4 \cdot 10^{-3}$.  For each model, we attacked vehicle pixels by perturbing the entire image and evaluated
the mean error caused by the global attacks (clear colors) and
the consistent global attacks (solid colors) over the corresponding
category.}
\label{fig:KITTI12_global}
\end{figure}

 \begin{figure}[th]
\begin{center}
   \includegraphics[width=0.99\linewidth]{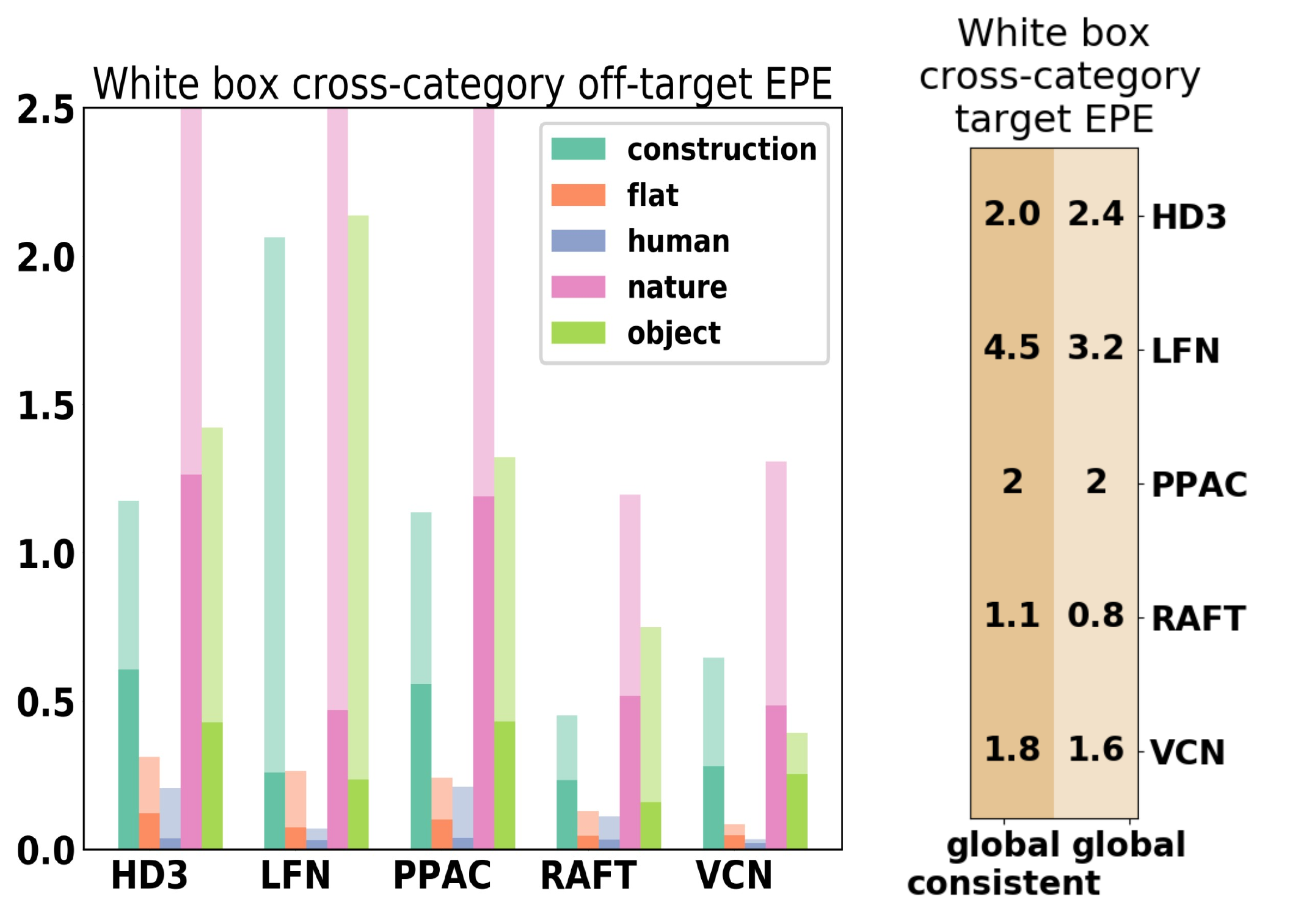}
\end{center}
   \caption{ cross category attacks on vehicles in the KITTI 12' dataset with $|\Delta I|=2 \cdot 10^{-2}$.  For each model, we attacked vehicle pixels by perturbing nature category pixels and evaluated the mean error caused by the global attacks (clear colors) and the consistent global attacks (solid colors) over the corresponding category.}
\label{fig:KITTI12_cross}
\end{figure}

\begin{figure}[th]
\begin{center}
   \includegraphics[width=0.99\linewidth]{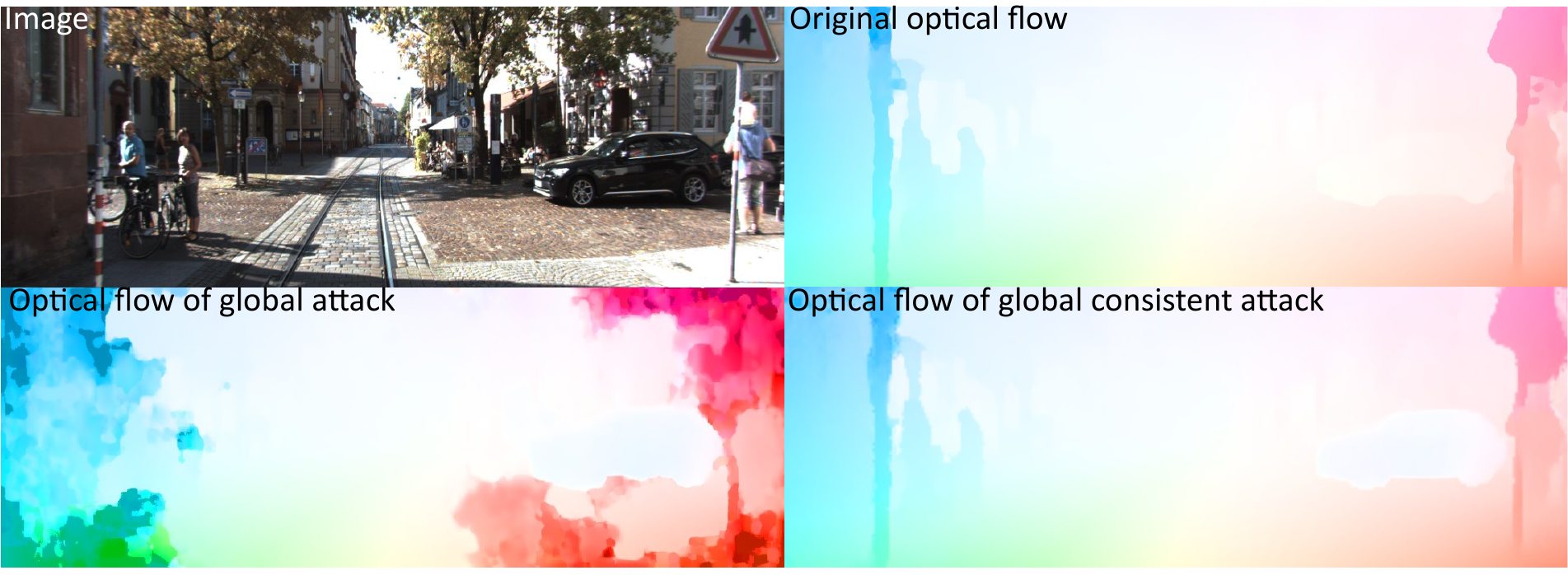}
\end{center}
   \caption{ A visualization of a global attack baseline and the effect of adding a consistency term on a vehicle instance using PPAC with $|\Delta{I}|= 4 \cdot  10^{-3}$. In the non-consistent case (second row) we see that the remaining scene is affected as well. Once a consistency term is being added (last row) the attack is much more focused on the attacked instance }
\label{fig:FigurePPACKITTI12}
\end{figure}
In the local setting we've perturbed vehicle pixels and evaluated our attack on vehicle pixels as well. Figure \ref{fig:KITTI12_local} presents the results of this experiment. We see that the off-target EPE was reduced in the consistency attack. Moreover, the on-target EPE varied only slightly by this addition.

For the global setting we've perturbed the entire image pixels to attack vehicle pixels. Figure \ref{fig:KITTI12_global} visualizes the result of the global experiment. As for the KITTI 15' case, we see that adding consistency reduces off-target EPE and increases on-target EPE. 

Finally, for the cross-category setting we've perturbed nature pixels to attack vehicle pixels. Figure \ref{fig:KITTI12_cross} visualizes the results of this experiment. We notice three main aspects of this experiment. First, the off-target EPE on all category was reduced. Second, the nature category that was perturbed gained the most adding a consistency term to the attack. Last, we see that the on-target efficiency has remained similar once adding the consistency term.

Figure \ref{fig:FigurePPACKITTI12} visualizes the effect of adding consistency for the KITTI 12' dataset attacks. In both the attacked flows (bottom row) we note that the car flow has changed. In the non-consistent attack (bottom left) we see that the remaining scene flow has changed as well. For the consistent case (bottom right), the only visible change is that of the car.

\begin{figure}[th!]
\begin{center}
\includegraphics[width=0.99\linewidth]{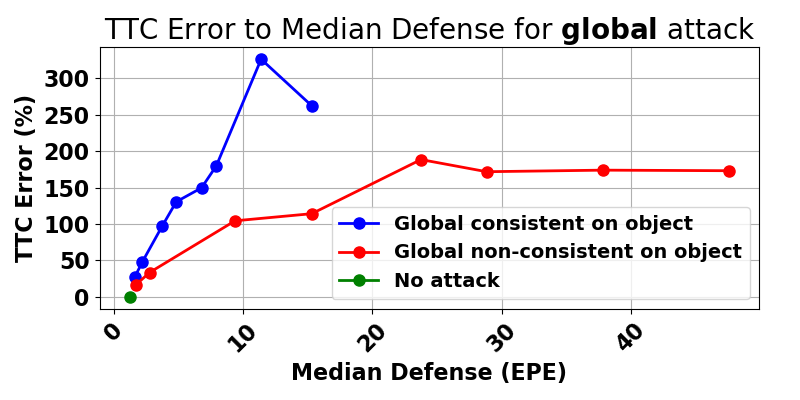} \\
\includegraphics[width=0.99\linewidth]{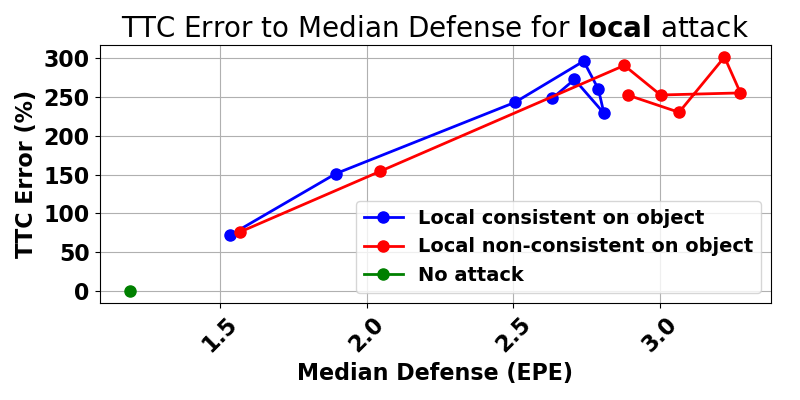}
\end{center}
\caption{TTC error to the Median AA detection score. The left and right sub-figures correspond to the global and local attacks, respectively. The Y-axis in all graphs corresponds to the TTC error, while the X-axis corresponds to the Median AA detection score. The graph is created using attacks with magnitude $||m\cdot 10^{-3}||$ for $m \in \{0.2, 0.4, 1.2, 2, 3.2, 4, 6, 8\}$.}
\label{fig:ttc_to_median}
\end{figure}

\subsection{TTC: The Median AA detection score}
Similarly to the results of the TTC as a function of the AA detection scores presented in Section 3.3 of the paper, we present the Median AA detection score, as a function of the TTC error. The graphs for the TTC error as a function of the Median detection score is presented in Figure~\ref{fig:ttc_to_median}. The results and trends for the TTC error as a function of the Median score are very similar to that of the warping error and the Gaussian AA detection scores.

\subsection{ KITTI 15' attack visualizations }

In this section we provide further visualizations of the vehicle attack conducted on the KITTI 15' dataset. Example of the local and global attack results are visualized. Each example include the first image of the optical flow pair, $I_{1}$, the original flow, consistent flow, and the non-consistent flow.
\vspace{10cm}

\begin{figure}[th]
\begin{center}
   \includegraphics[width=0.99\linewidth]{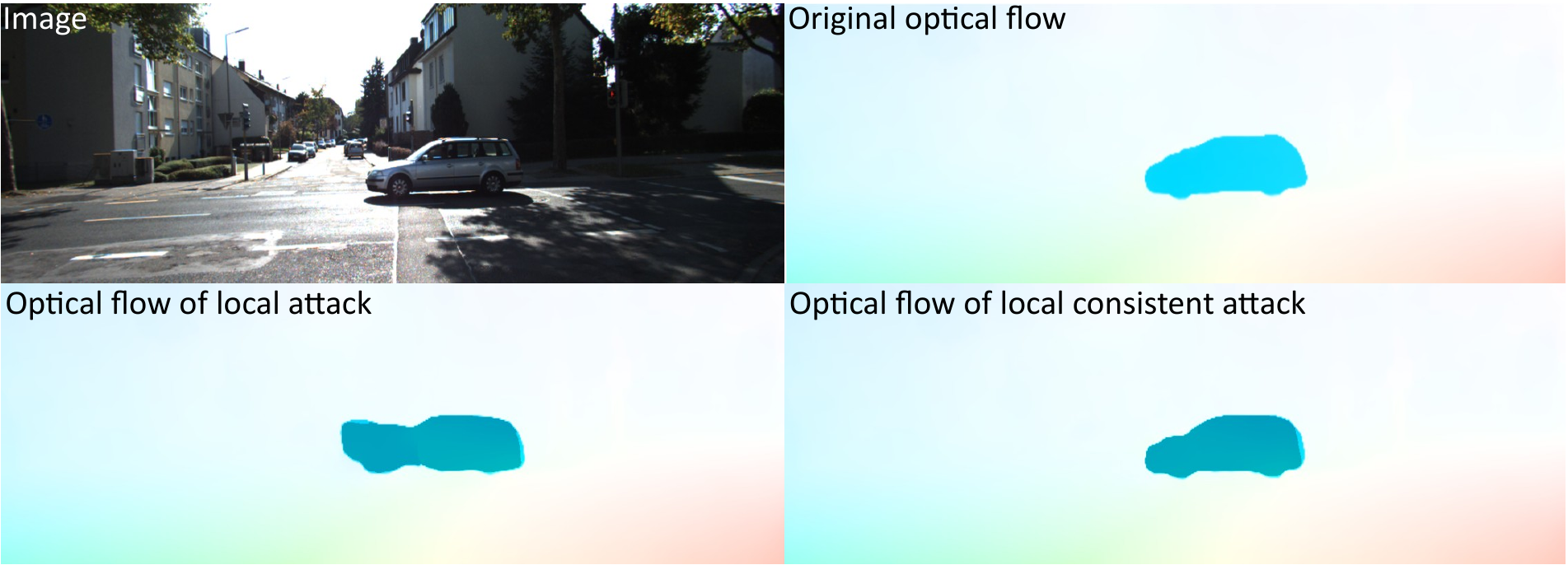}
\end{center}
   \caption{ A visualization of a local attack baseline and the effect of adding a consistency term on a vehicle instance using LFN with $|\Delta{I}|= 4 \cdot  10^{-3}$. Note that vehicle outline changes. In the non-consistent case (second row) we see that the remaining scene is affected as well, and once a consistency term is being added (last row) the attack is much more focused on the attacked instance }
\label{fig:FigureAppendix_1}
\end{figure}

\begin{figure}[th]
\begin{center}
   \includegraphics[width=0.99\linewidth]{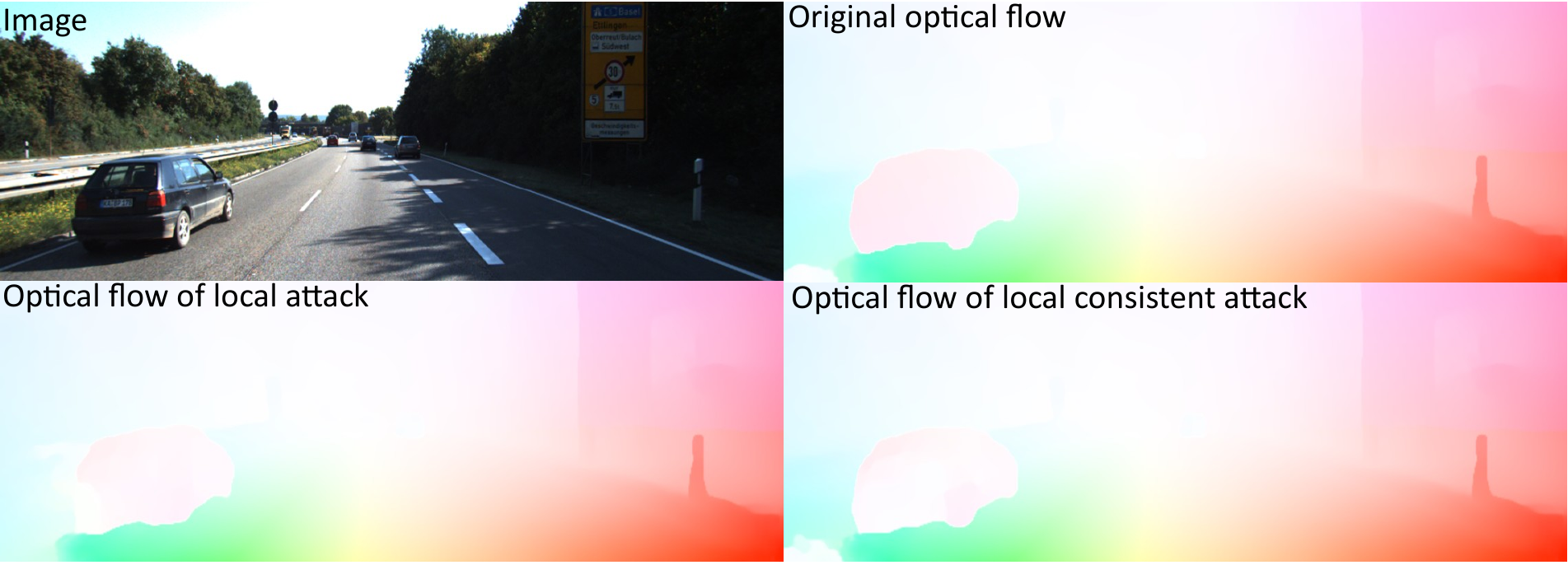}
\end{center}
   \caption{ A visualization of a local attack baseline and the effect of adding a consistency term on a vehicle instance using LFN with $|\Delta{I}|= 4 \cdot  10^{-3}$. The road behind the vehicle is effected from the attack. In the non-consistent case (second row) we see that the remaining scene is affected as well, and once a consistency term is being added (last row) the attack is much more focused on the attacked instance }
\label{fig:FigureAppendix_2}
\end{figure}

\begin{figure}[t]
\begin{center}
   \includegraphics[width=0.99\linewidth]{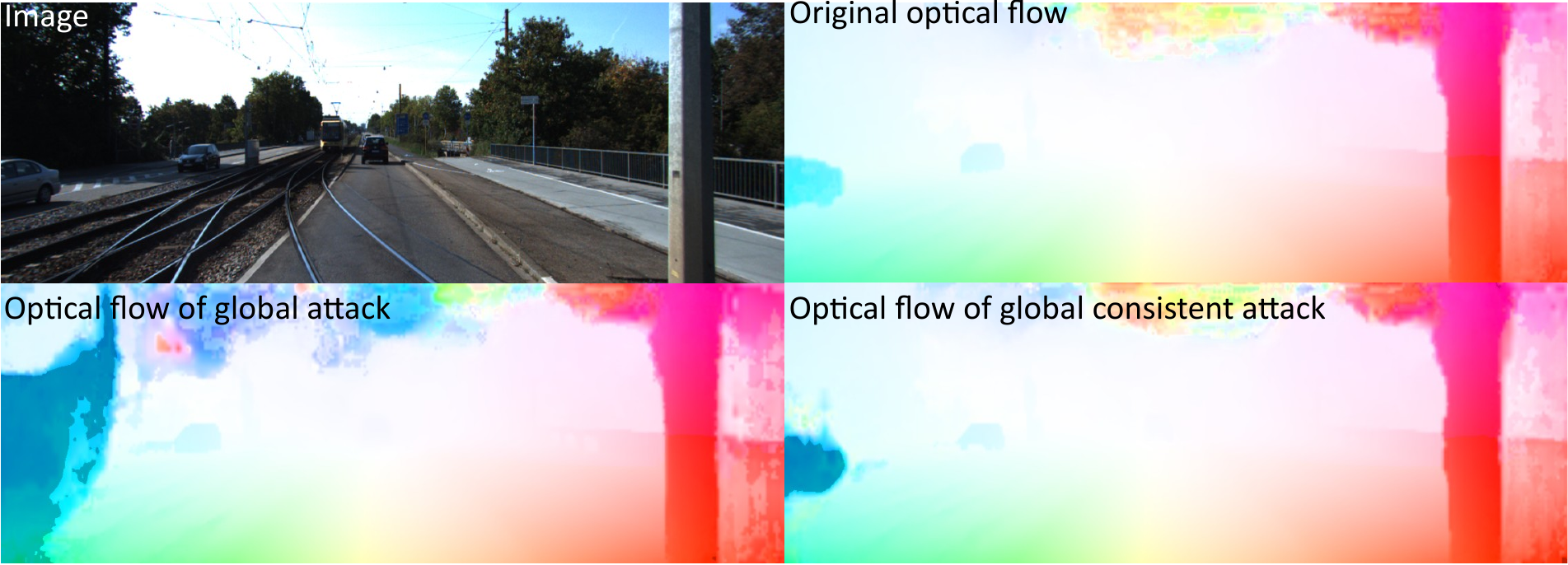}
\end{center}
   \caption{ A visualization of a global attack baseline and the effect of adding a consistency term on a vehicle instance using HD3 with $|\Delta{I}|= 4 \cdot  10^{-3}$. In the non-consistent case (second row) we see that the remaining scene is affected as well, and once a consistency term is being added (last row) the attack is much more focused on the attacked instance }
\label{fig:FigureAppendix_3}
\end{figure}

\begin{figure}[t]
\begin{center}
   \includegraphics[width=0.99\linewidth]{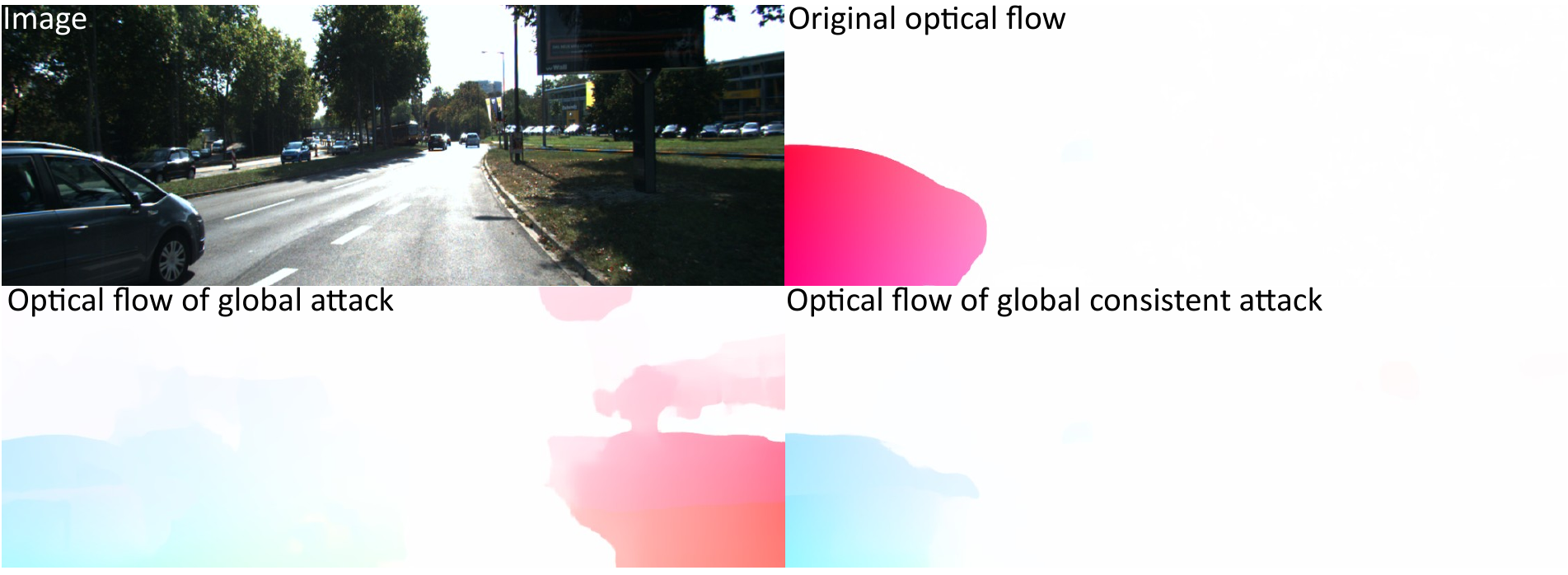}
\end{center}
   \caption{ A visualization of a global attack baseline and the effect of adding a consistency term on a vehicle instance using RAFT with $|\Delta{I}|= 4 \cdot  10^{-2}$. In the non-consistent case (second row) we see that the remaining scene is affected as well, and once a consistency term is being added (last row) the attack is much more focused on the attacked instance }
\label{fig:RAFT_1}
\end{figure}
\begin{figure}[t]
\begin{center}
   \includegraphics[width=0.99\linewidth]{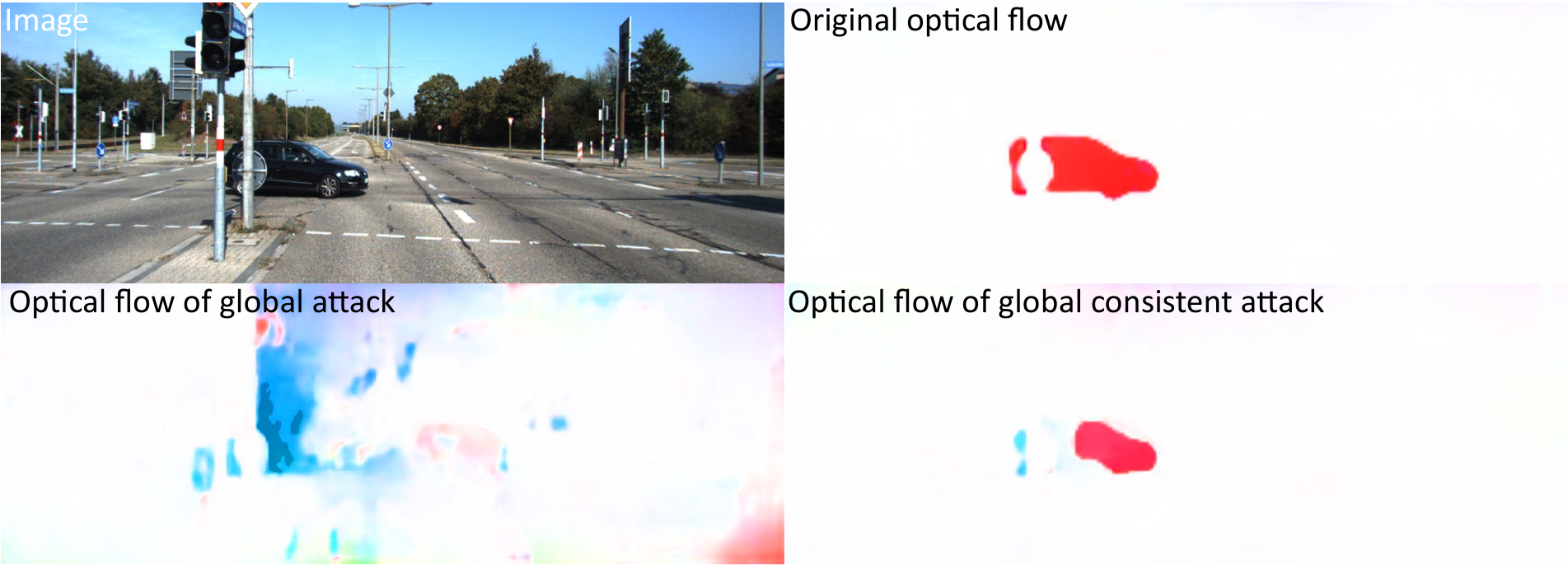}
\end{center}
   \caption{ A visualization of a global attack baseline and the effect of adding a consistency term on a vehicle instance using VCN with $|\Delta{I}|= 4 \cdot  10^{-2}$. In the non-consistent case (second row) we see that the remaining scene is affected as well, and once a consistency term is being added (last row) the attack is much more focused on the attacked instance }
\label{fig:VCN_1}
\end{figure}
\vspace{0px}
\begin{figure}[b!]
\begin{center}
   \includegraphics[width=0.99\linewidth]{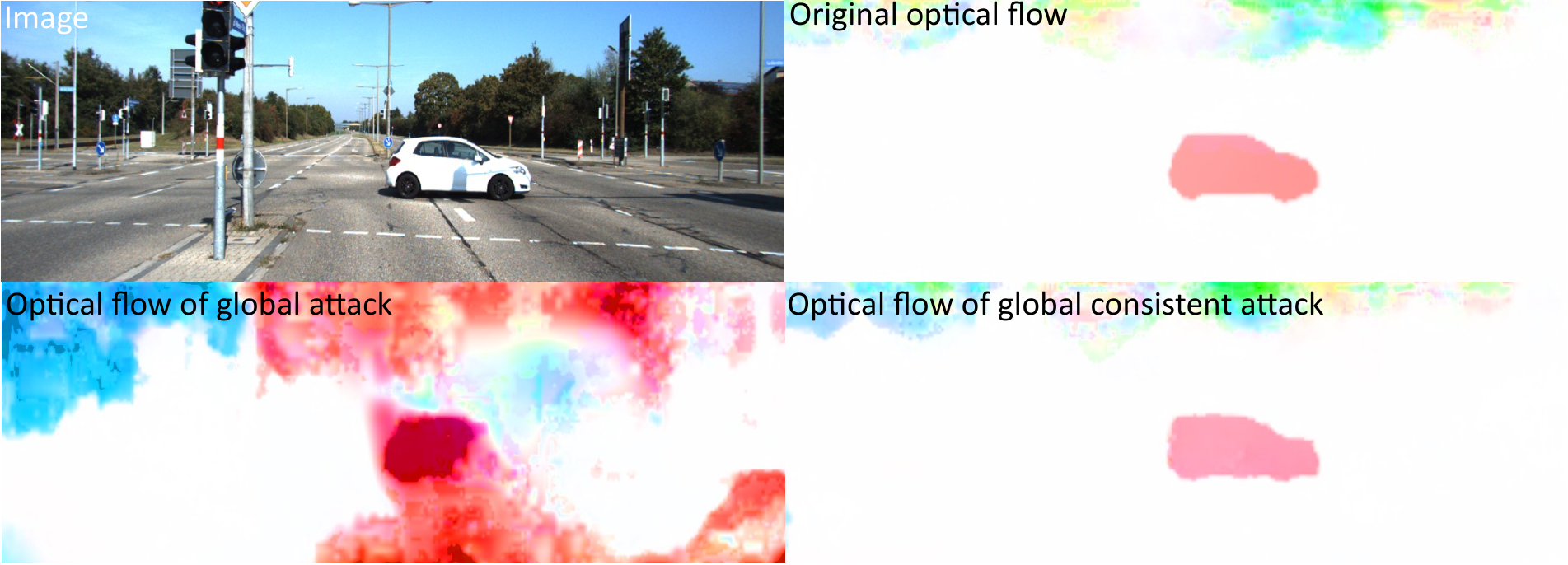}
\end{center}
   \caption{ A visualization of a global attack baseline and the effect of adding a consistency term on a vehicle instance using HD3 with $|\Delta{I}|= 2 \cdot 10^{-2}$.  In the non-consistent case (second row) we see that the remaining scene is affected as well, and once a consistency term is being added (last row) the attack is much more focused on the attacked instance }
\label{fig:FigureAppendix_4}
\end{figure}

\clearpage
{\small
\bibliographystyle{ieee_fullname}
\bibliography{targetted_aa}
}

\end{document}